\documentclass[10pt,conference,a4paper]{IEEEtran}

\usepackage{ifpdf}

\usepackage{cite}

\ifCLASSINFOpdf
   \usepackage[pdftex]{graphicx}
\else
\fi
\ifCLASSOPTIONcompsoc
 \usepackage[caption=false,font=normalsize,labelfont=sf,textfont=sf]{subfig}
\else
 \usepackage[caption=false,font=footnotesize]{subfig}
\fi
\usepackage{url}

\usepackage{amsmath} 
\usepackage{amssymb}  

\usepackage{algorithm}
\usepackage{algorithmicx}
\usepackage{algpseudocode}
\usepackage[linktocpage=true]{hyperref}
\hypersetup{
    colorlinks=true,
    linkcolor=blue,
    filecolor=magenta,      
    urlcolor=cyan,
}

\newcommand{\etal}{\textit{et al}. }

\usepackage[export]{adjustbox}

\usepackage[flushleft]{threeparttable}

\usepackage{tikz}
\usetikzlibrary{spy,calc}

\newif\ifblackandwhitecycle
\gdef\patternnumber{0}

\pgfkeys{/tikz/.cd,
    zoombox paths/.style={
        draw=orange,
        very thick
    },
    black and white/.is choice,
    black and white/.default=static,
    black and white/static/.style={ 
        draw=white,   
        zoombox paths/.append style={
            draw=white,
            postaction={
                draw=black,
                loosely dashed
            }
        }
    },
    black and white/static/.code={
        \gdef\patternnumber{1}
    },
    black and white/cycle/.code={
        \blackandwhitecycletrue
        \gdef\patternnumber{1}
    },
    black and white pattern/.is choice,
    black and white pattern/0/.style={},
    black and white pattern/1/.style={    
            draw=white,
            postaction={
                draw=black,
                dash pattern=on 2pt off 2pt
            }
    },
    black and white pattern/2/.style={    
            draw=white,
            postaction={
                draw=black,
                dash pattern=on 4pt off 4pt
            }
    },
    black and white pattern/3/.style={    
            draw=white,
            postaction={
                draw=black,
                dash pattern=on 4pt off 4pt on 1pt off 4pt
            }
    },
    black and white pattern/4/.style={    
            draw=white,
            postaction={
                draw=black,
                dash pattern=on 4pt off 2pt on 2 pt off 2pt on 2 pt off 2pt
            }
    },
    zoomboxarray inner gap/.initial=5pt,
    zoomboxarray columns/.initial=2,
    zoomboxarray rows/.initial=2,
    subfigurename/.initial={},
    figurename/.initial={zoombox},
    zoomboxarray/.style={
        execute at begin picture={
            \begin{scope}[
                spy using outlines={%
                    zoombox paths,
                    width=\imagewidth / \pgfkeysvalueof{/tikz/zoomboxarray columns} - (\pgfkeysvalueof{/tikz/zoomboxarray columns} - 1) / \pgfkeysvalueof{/tikz/zoomboxarray columns} * \pgfkeysvalueof{/tikz/zoomboxarray inner gap} -\pgflinewidth,
                    height=\imageheight / \pgfkeysvalueof{/tikz/zoomboxarray rows} - (\pgfkeysvalueof{/tikz/zoomboxarray rows} - 1) / \pgfkeysvalueof{/tikz/zoomboxarray rows} * \pgfkeysvalueof{/tikz/zoomboxarray inner gap}-\pgflinewidth,
                    magnification=3,
                    every spy on node/.style={
                        zoombox paths
                    },
                    every spy in node/.style={
                        zoombox paths
                    }
                }
            ]
        },
    execute at end picture={
            \end{scope}
            \node at (image.north) [anchor=north,inner sep=0pt] {{\phantomimage}};
            \node at (zoomboxes container.north) [anchor=north,inner sep=0pt] {{\phantomimage}};
     \gdef\patternnumber{0}
        },
        spymargin/.initial=0.5em,
        zoomboxes xshift/.initial=1,
        zoomboxes right/.code=\pgfkeys{/tikz/zoomboxes xshift=1},
        zoomboxes left/.code=\pgfkeys{/tikz/zoomboxes xshift=-1},
        zoomboxes yshift/.initial=0,
        zoomboxes above/.code={
            \pgfkeys{/tikz/zoomboxes yshift=1},
            \pgfkeys{/tikz/zoomboxes xshift=0}
        },
        zoomboxes below/.code={
            \pgfkeys{/tikz/zoomboxes yshift=-1},
            \pgfkeys{/tikz/zoomboxes xshift=0}
        },
        caption margin/.initial=4ex,
    },
    adjust caption spacing/.code={},
    image container/.style={
        inner sep=0pt,
        at=(image.north),
        anchor=north,
        adjust caption spacing
    },
    zoomboxes container/.style={
        inner sep=0pt,
        at=(image.north),
        anchor=north,
        name=zoomboxes container,
        xshift=\pgfkeysvalueof{/tikz/zoomboxes xshift}*(\imagewidth+\pgfkeysvalueof{/tikz/spymargin}),
        yshift=\pgfkeysvalueof{/tikz/zoomboxes yshift}*(\imageheight+\pgfkeysvalueof{/tikz/spymargin}+\pgfkeysvalueof{/tikz/caption margin}),
        adjust caption spacing
    },
    calculate dimensions/.code={
        \pgfpointdiff{\pgfpointanchor{image}{south west} }{\pgfpointanchor{image}{north east} }
        \pgfgetlastxy{\imagewidth}{\imageheight}
        \global\let\imagewidth=\imagewidth
        \global\let\imageheight=\imageheight
        \gdef\columncount{1}
        \gdef\rowcount{1}
        
    },
    image node/.style={
        inner sep=0pt,
        name=image,
        anchor=south west,
        append after command={
            [calculate dimensions]
            node [image container,subfigurename=\pgfkeysvalueof{/tikz/figurename}-image] {\phantomimage}
            node [zoomboxes container,subfigurename=\pgfkeysvalueof{/tikz/figurename}-zoom] {\phantomimage}
        }
    },
    color code/.style={
        zoombox paths/.append style={draw=#1}
    },
    connect zoomboxes/.style={
    spy connection path={\draw[draw=none,zoombox paths] (tikzspyonnode) -- (tikzspyinnode);}
    },
    help grid code/.code={
        \begin{scope}[
                x={(image.south east)},
                y={(image.north west)},
                font=\footnotesize,
                help lines,
                overlay
            ]
            \foreach \x in {0,1,...,9} { 
                \draw(\x/10,0) -- (\x/10,1);
                \node [anchor=north] at (\x/10,0) {0.\x};
            }
            \foreach \y in {0,1,...,9} {
                \draw(0,\y/10) -- (1,\y/10);                        \node [anchor=east] at (0,\y/10) {0.\y};
            }
        \end{scope}    
    },
    help grid/.style={
        append after command={
            [help grid code]
        }
    },
}

\newcommand\phantomimage{%
    \phantom{%
        \rule{\imagewidth}{\imageheight}%
    }%
}
\newcommand\zoombox[2][]{
    \begin{scope}[zoombox paths]
        \pgfmathsetmacro\xpos{
            (\columncount-1)*(\imagewidth / \pgfkeysvalueof{/tikz/zoomboxarray columns} + \pgfkeysvalueof{/tikz/zoomboxarray inner gap} / \pgfkeysvalueof{/tikz/zoomboxarray columns} ) + \pgflinewidth
        }
        \pgfmathsetmacro\ypos{
            (\rowcount-1)*( \imageheight / \pgfkeysvalueof{/tikz/zoomboxarray rows} + \pgfkeysvalueof{/tikz/zoomboxarray inner gap} / \pgfkeysvalueof{/tikz/zoomboxarray rows} ) + 0.5*\pgflinewidth
        }
        \edef\dospy{\noexpand\spy [
            #1,
            zoombox paths/.append style={
                black and white pattern=\patternnumber
            },
            every spy on node/.append style={#1},
            x=\imagewidth,
            y=\imageheight
        ] on (#2) in node [anchor=north west] at ($(zoomboxes container.north west)+(\xpos pt,-\ypos pt)$);}
        \dospy
        \pgfmathtruncatemacro\pgfmathresult{ifthenelse(\columncount==\pgfkeysvalueof{/tikz/zoomboxarray columns},\rowcount+1,\rowcount)}
        \global\let\rowcount=\pgfmathresult
        \pgfmathtruncatemacro\pgfmathresult{ifthenelse(\columncount==\pgfkeysvalueof{/tikz/zoomboxarray columns},1,\columncount+1)}
        \global\let\columncount=\pgfmathresult
        \ifblackandwhitecycle
            \pgfmathtruncatemacro{\newpatternnumber}{\patternnumber+1}
            \global\edef\patternnumber{\newpatternnumber}
        \fi
    \end{scope}
}

\begin{document}
%
\title{Incremental 3D Line Segment Extraction from Semi-dense SLAM}

\author{\IEEEauthorblockN{Shida He \qquad\qquad Xuebin Qin \qquad\qquad Zichen Zhang \qquad\qquad Martin Jagersand}
\IEEEauthorblockA{University of Alberta}}

\maketitle

\begin{abstract}

Although semi-dense Simultaneous Localization and Mapping (SLAM) has been becoming more popular over the last few years, there is a lack of efficient methods for representing and processing their large scale point clouds.
In this paper, we propose using 3D line segments to simplify the point clouds generated by semi-dense SLAM. 
Specifically, we present a novel incremental approach for 3D line segment extraction. 
This approach reduces a 3D line segment fitting problem into two 2D line segment fitting problems and takes advantage of both images and depth maps.
In our method, 3D line segments are fitted incrementally along detected edge segments via minimizing fitting errors on two planes.
By clustering the detected line segments, the resulting 3D representation of the scene achieves a good balance between compactness and completeness. 
Our experimental results show that the 3D line segments generated by our method are highly accurate.
As an application, we demonstrate that these line segments greatly improve the quality of 3D surface reconstruction compared to a feature point based baseline.

\end{abstract}


%
\IEEEpeerreviewmaketitle

\section{Introduction}


Reconstructing surface from a stream of images is an important yet challenging topic in computer vision.
The 3D surface of the scene is essential for a variety of applications, such as photogrammetry, robot navigation, augmented reality, etc.

Given the expanding point cloud from a SLAM system, incremental surface reconstruction algorithms can be used to reconstruct a 3D surface representing the scene \cite{lovi2011incremental, hoppe2013incremental}. 
With the recent development in SLAM, some monocular SLAM systems can produce dense or semi-dense point cloud which represents the scene more completely than the sparse point cloud produced by traditional SLAMs \cite{mur2015probabilistic, engel2014lsd}. 
Although the surface reconstruction algorithms can run in real-time on sparse point clouds, the statement does not hold true when the number of points is increased significantly.
In this paper, we propose to represent the scene using line segments in order to simplify the result of semi-dense SLAM and achieve real-time performance in the task of surface reconstruction.

Line segments efficiently preserve more structural information of a scene than point clouds.
Many line segment based 3D reconstruction algorithms have been proposed \cite{DBLP:journals/cviu/BartoliS05, line3D, pumarola2017pl, DBLP:journals/ijcv/MicusikW17, snow2016line, DBLP:conf/accv/NakayamaSSY14}.
Most of these methods rely on efficient line segment detection and inter-keyframe matching, which are difficult for certain scenes.
However, with the dense or semi-dense point clouds available, one can estimate 3D line segments without explicit matching and triangulation.

In this paper, we develop a novel method to incrementally extract 3D line segments from semi-dense SLAM. 
Those detected line segments present structural relations among points and prove to be a highly efficient form of 3D representation. 
Our method focuses on the accuracy of reconstructed line segments instead of the improvement to camera localization in other works \cite{pumarola2017pl, DBLP:conf/accv/NakayamaSSY14}.
Specifically, we propose a novel edge aided 3D line segment fitting algorithm.
We directly detect 3D line segments by taking both 2D locations and corresponding depth into consideration in the line segment detection procedure.
Our method produces accurate 3D line segments with few outliers.
We utilize the 3D line segments extracted by our method in incremental surface reconstruction and improve the quality of reconstructed surface with respect to a feature point based baseline.

The rest of this paper is organized as follows. 
In Section 2, we review some related works about 3D line segment reconstruction. 
In Section 3, we describe our proposed algorithm in details. 
We present our experimental results in Section 4 and conclude in Section 5.

\section{Related Works}

Existing 3D line segment detection methods can be categorized into three main classes:

\subsubsection{3D line reconstruction from matching 2D lines}
Bartoli and Sturm proposed a line segment based Structure from Motion (SFM) pipeline \cite{DBLP:journals/cviu/BartoliS05}. 
Micusik and Wilder used relaxed endpoint constraints for line matching and developed a SLAM-like line segment based SFM system \cite{DBLP:journals/ijcv/MicusikW17}. 
Hofer \etal matched 2D line segments detected from images of different views by pure geometric constraints \cite{line3D}. 
In \cite{pumarola2017pl}, Pumarola \etal proposed PLSLAM, which detect line segments using LSD \cite{von2010lsd} and match them based on LBD descriptor \cite{zhang2013efficient}. 

\subsubsection{3D line fitting directly from 3D points}
Roberts proposed a compact representation for a line in \cite{DBLP:conf/cvpr/Roberts88}.
Using this representation, Snow and Schaffrin developed an algorithm for solving the Total Least-Squares problem of 3D line fitting \cite{snow2016line}.
However, this kind of methods are sensitive to noise and outliers. 
Random Sample Consensus (RANSAC) is relatively robust to small number of outliers \cite{RANSAC}. 
However, RANSAC is time consuming in 3D space and the optimal line fitting is not guaranteed in the presence of a large amount of outliers.

\subsubsection{3D line extraction aided by 2D cues}
Woo \etal \cite{DBLP:conf/pcm/WooHJL05} detected 2D line segments from 2D aerial images first, and then used their corresponding 3D points on buildings' Digital Elevation Model (DEM) to fit 3D lines. 
Given RGB-D sensor, Nakayama \etal \cite{DBLP:conf/accv/NakayamaSSY14} transformed 2D points on detected 2D line segments directly to 3D using corresponding depth image. Then, 3D line segments are fitted by RANSAC in 3D space. 

Since SLAM systems can output semi-dense point clouds as their mapping results, we prefer to make full use of these results and extract 3D line segments using both image and point cloud information. 
Our method belongs to the third class because we extract line segments from the semi-dense point cloud with the help of detected 2D edges.  

\begin{figure}[tbp]
\begin{center}
   \includegraphics[width=0.6\linewidth]{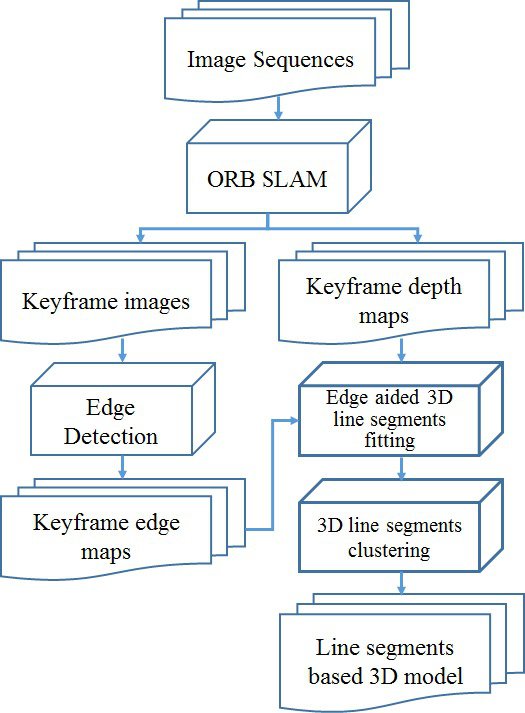}
\end{center}
   \caption{Workflow of our method. The input of our method is a video (image sequence) which is captured by a calibrated moving camera. The output is a line segment based 3D model of the scene.}
\label{fig:workflow}
\end{figure}


\section{Method}

Instead of detecting 2D line segments and finding their corresponding 3D points for fitting 3D line segments afterwards, we fit the line segment in 3D by iteratively fitting its projections in two different planes.

Specifically, in order to extract 3D line segments from semi-dense point cloud, our method performs the following steps on each new keyframe (shown in Figure \ref{fig:workflow}): (1)
Compute semi-dense depth map; (2) Edge aided 3D line segment fitting; (3) 3D line segment clustering and filtering.


\subsection{Keyframe and depth map generation}

Our method is based on ORB-SLAM \cite{ORB} with semi-dense module \cite{mur2015probabilistic}.
ORB-SLAM is a feature based SLAM system which takes the image sequence from a moving camera and computes the camera poses in real time.
Mur-Artal and Tard{\'{o}}s \cite{mur2015probabilistic} presented a semi-dense module which is able to compute a semi-dense depth map for each keyframe.
In principle, other keyframe based dense or semi-dense SLAM system could be used to generate the semi-dense depth maps, such as LSD-SLAM \cite{engel2014lsd}.

\begin{figure}[t]
\begin{center}
   \includegraphics[width=0.5\linewidth]{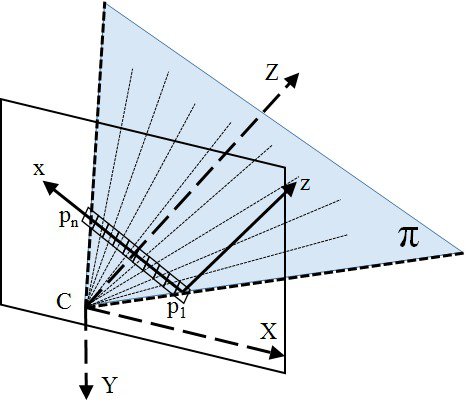}
\end{center}
   \caption{Line segment fitting related coordinates.
   $\textbf{C}$-$\textbf{XYZ}$ is the camera coordinates. The X-axis and Y-axis are parallel to the image coordinates. The Z-axis is the depth direction. $\{ p_1... p_n\}$ represent a detected 2D image line segment. Line segment pixels $\{ p_1...p_n\}$ and their corresponding real-world points are all located on the same plane $\pi$. The x-axis is defined by vector $\protect\overrightarrow{ p_1 p_n}$ while z-axis is parallel to the Z-axis. For each 3D line segment, we fit two 2D lines in image coordinate plane and $\bf{p_1}$-$\textbf{xz}$ coordinate plane.}
\label{figCoordinates}
\end{figure}

\subsection{Edge aided 3D line segment fitting}

Fitting 3D line segments directly from point clouds can be difficult and time consuming. 
In this paper, we propose to use 2D image information on keyframes to help 3D line segment fitting from semi-dense point clouds.

We first extract edge segments from keyframes using Edge Drawing \cite{EdgeDrawing}. 
Edge Drawing is a linear time edge detector which can produce a set of accurate, contiguous and clean edge segments represented by one-pixel-wide pixel chains.
Now the detected edge segments of the a keyframe can be expressed as 
$ ES = \{es_1,es_2, ..., es_n\}$ where $ es_i = \{p_1, p_2, ..., p_m\}$ is an edge segment formulated as a pixel chain. 
$ p_i$ represents a pixel which is a vector of $(x,y,Z)$ where $x$ and $y$ are the image coordinates and $Z$ is its corresponding depth in the semi-dense depth map. 
The number of edge segments in a keyframe and number of pixels in an edge segment are denoted by $n$ and $m$ respectively. 
It is worth noting that image pixels with high intensity gradient are more likely to be selected for computing depth value in the semi-dense SLAM system. 
Edge segments are detected based on pixel intensity gradients as well.
Thus, the detected edge pixels are very likely to have depth values. 
The pixels which have no depth values will be considered as outliers in the line fitting process.

\begin{algorithm}[tbp]
	\caption{Edge aided 3D line segment fitting} 
	{\bf Input:}
	An edge segment which is a list of pixels: $es = \{p_1, p_2, ..., p_m\}$, where $p_i$ denotes the $i$-th pixel on the edge segment \\
	{\bf Output:}
	A set of fitted 3D line segments: $LS$
	\begin{algorithmic}[1]
	\State Initialize two empty set of pixels: $pixels$, $outliers$
	\For{\textbf{each} $p_i \in es$}
	\If{$pixels = \varnothing$}
	\State Move first $L$ pixels in $es$ to $pixels$
	\State Fit two 2D lines $l_{im}$ and $l_{depth}$ to $pixels$
	\EndIf
	\State Compute distance $d_{im}$ from $(p_i.x,p_i.y)$ to $l_{im}$ 
	\State Compute distance $d_{depth}$ from $(p_i.D,p_i.Z)$ to $l_{depth}$
	\If{$d_{im} < e_1 \ \& \ d_{depth} < e_2$}
	\State Move $p_i$ to $pixels$ 
	\Else 
	\State Move $p_i$ to $outliers$ 
	\EndIf
	\If{$es = \varnothing \parallel \left\vert{outliers}\right\vert > L$}
	\If{$\left\vert{pixels}\right\vert > L$}
	\State Fit $l_{im}$ and $l_{depth}$ to $pixels$
	\State Compute the 3D line segment and add to $LS$
	\EndIf
	\State Empty $pixles$ and $outliers$
	\EndIf
	\EndFor
	
		
		
		
		
		
		
	\end{algorithmic}
	\label{alg:LineFitting}
\end{algorithm}

3D line segments of a keyframe are extracted from those detected image edge segments by Algorithm \ref{alg:LineFitting}. 
The main idea of this algorithm is to reduce a 3D line fitting problem to two 2D line fitting problems. 
The coordinate frames are defined as shown in Figure \ref{figCoordinates}.
For each edge segment, the algorithm initially takes its first $L$ pixels to fit two 2D lines ($l_{im}$ and $l_{depth}$) in the image coordinate frame and the $\bf{p_1}$-$\textbf{xz}$ coordinate frame using total least square method. 
The line $l_{im}$ is fitted based on the pixels' $(x,y)$ values while $l_{depth}$ is fitted based on $(D,Z)$.
$Z$ is the pixel's depth and $D$ is the distance from $ p_1$ to the pixel's projection on the x-axis.
Total least square 2D line fitting is performed by solving Singular Value Decomposition (SVD) \cite{golub1980analysis}.
It is worth noting that the plane $\bf{p_1}$-$\textbf{xz}$ is not always the same with plane $\pi$, which is determined by $C$, $p_1$ and $p_n$.
The plane $\bf{p_1}$-$\textbf{xz}$ is orthogonal to the image plane where $l_{im}$ is fitted. The two planes, $\bf{p_1}$-$\textbf{xz}$ and $\pi$, are the same plane only if the image center lies on the line segment $p_1p_n$. 
Although the 3D points may not be located on the plane $\bf{p_1}$-$\textbf{xz}$, it is much easier to perform line fitting on $\bf{p_1}$-$\textbf{xz}$ rather than $\pi$. 
Given the image plane coordinate of the points, actual 3D positions can be easily recovered using the depth of points in $\bf{p_1}$-$\textbf{xz}$ plane.

After obtain an initial line segment, we compute the distances of the next pixel in pixel chain to $l_{im}$ and $l_{depth}$ in their corresponding coordinate frames. 
Note that $D$ and $Z$ have different units. 
To have the same unit as $D$, $Z$ is multiplied by the focal length $f$ before distance computation.
If both distances are smaller than certain thresholds, $e_1$ and $e_2$ respectively, we will add the pixel to the fitted pixel set to extend the line. 
Otherwise the pixel will be considered as an outlier. 
If $L$ consecutive pixels are outliers, we stop the current line segment search and start a new line segment search. 
Another pair of total least square fittings on the two planes are performed to obtain the final 3D line for each line segment.
The 3D line segments are represented by their endpoints, which are estimated by projecting the points corresponding to its first and last pixel on to the final 3D line.
After traversing all of the edge segments of the keyframes, we can aggregate one 3D line segment set $ LS_k$ for each keyframe.

\subsection{3D line segment clustering and filtering}

To obtain a consistent reconstruction of the environment, 3D line segments $LS_{all} = \{LS_1, LS_2, ..., LS_n\}$ extracted from different keyframes are first registered to the same world coordinate system. 
The registered 3D line segments are denoted as $ls_{all} = \{ls_1, ls_2, ..., ls_w\}$. 
Here $w$ denotes the total number of 3D line segments from all keyframes. 
Directly registering all 3D line segments will produce redundant and slightly misaligned result.
We address this problem by proposing a simple incremental merging method. 

\begin{figure}
	\begin{center}
		\includegraphics[width=0.45\linewidth]{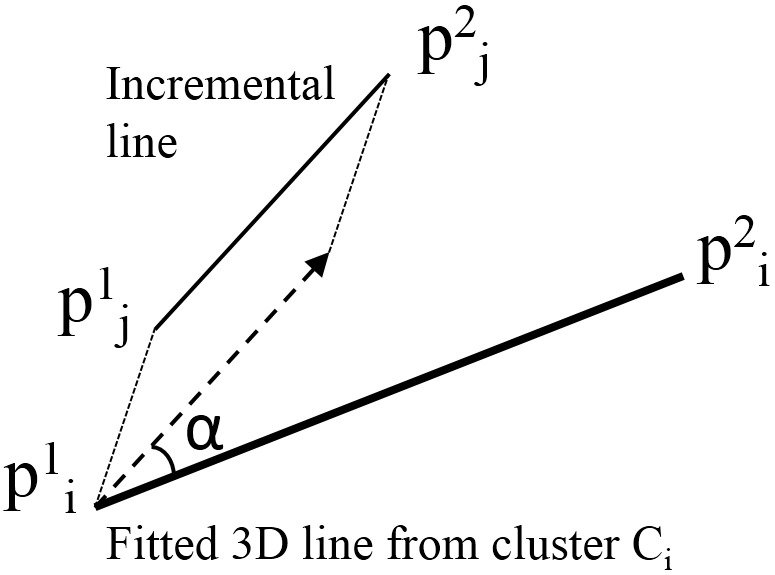}
		\includegraphics[width=0.45\linewidth]{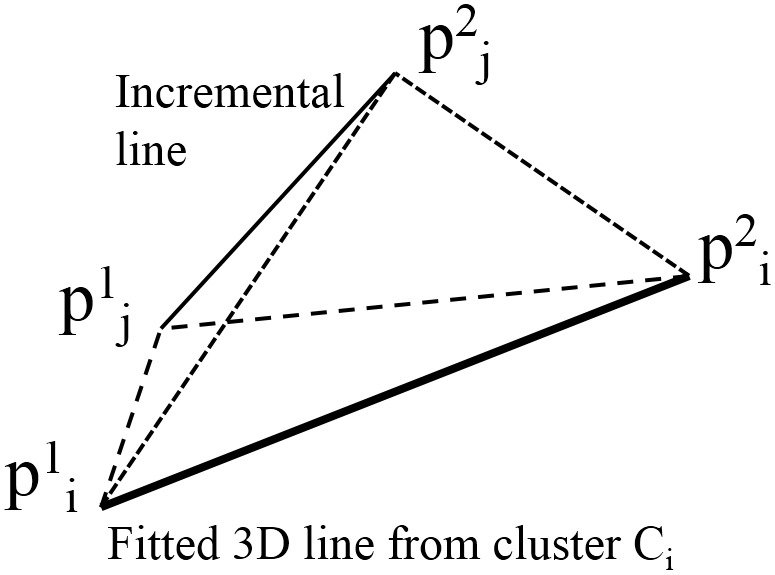}
		\caption{Clustering by angle and distance. $p^1_ip^2_i$ is fitted from an existing 3D line segment cluster, while $p^1_jp^2_j$ is an unclustered 3D line segment.}
		\label{fig:LineCluster}
	\end{center}
\end{figure}

The main idea of our merging method is clustering closely located 3D line segments and fitting those cluster sets with new 3D line segments.
As illustrated in Figure \ref{fig:LineCluster}, the angle and distance measures are used for clustering. 
The angle measure $\alpha$ is defined as:
\begin{equation}
    \alpha=acos(\frac{\overrightarrow{p^1_jp^2_j}\cdot \overrightarrow{p^1_ip^2_i}}{|\overrightarrow{p^1_jp^2_j}||\overrightarrow{p^1_ip^2_i}|})
\end{equation}
The distance measure $d$ is computed as:
\begin{align}
    d &=min(d_1,d_2)\\
    d_1 &= |\overrightarrow{p^1_jp^1_i}|+|\overrightarrow{p^1_jp^2_i}|-|\overrightarrow{p^1_ip^2_i}|\\
    d_2 &= |\overrightarrow{p^2_jp^1_i}|+|\overrightarrow{p^2_jp^2_i}|-|\overrightarrow{p^1_ip^2_i}|   
\end{align}
Specifically, we take the first 3D line segment $ls_1$ as the initial cluster $C_1$.
Then, we compute the angle and distance measure between the initial cluster (single line segment) and the next 3D line segment $ls_2$. 
If the angle $\alpha$ and distance $d$ are smaller than certain thresholds ($\lambda_\alpha$ and $\lambda_d$ respectively), we add $ls_2$ to the cluster $C_1$.
Otherwise, we create a new cluster $C_2$. 
For each cluster, if it contains more than one 3D line segments, we will fit a new 3D line segment to represent the cluster. 
The direction of the new line segment is determined by performing SVD on the matrix consisting of all points in $P_{ep}$, where $P_{ep}$ denotes the set containing all the endpoints of line segments in this cluster.
A new 3D infinite line is then determined by the direction together with the centroid of $P_{ep}$. 
In order to obtain a 3D line segment from this infinite line, 
we project endpoints in $P_{ep}$ onto the generated infinite 3D line and compute the furthest projections with respect to the centroid in both directions.
The 3D line segment between these two furthest projection points is taken as the fitted line segment of the cluster. 
This process is repeated until all the line segments in $ls$ are clustered.
Clusters with small size ($\left\vert{C_i}\right\vert < \lambda_C$) are filtered out in the end. 
In this way, we can merge a large number of line segments into fewer clusters and generate new 3D line segments with higher quality.

\begin{figure*}
	\begin{center}
	
	    \subfloat[Sample image]{
		\centering
		\begin{minipage}[t]{0.15\textwidth}
        \centering
		\includegraphics[height=0.85\linewidth, width=\linewidth]{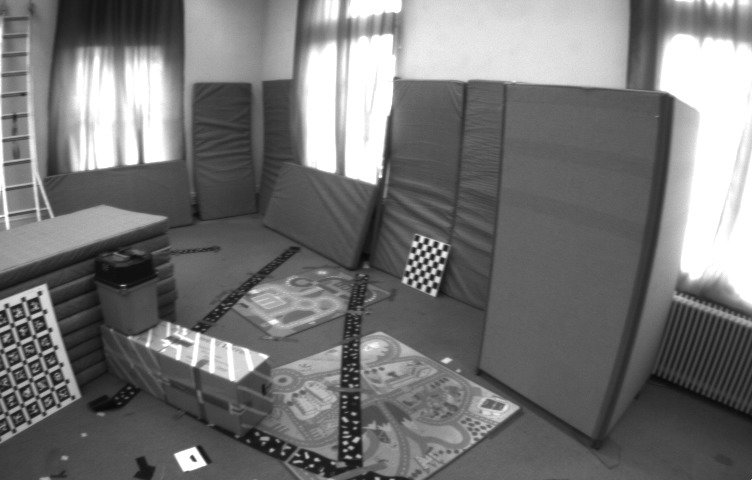}\\
		\vspace{0.05in}
		\includegraphics[height=0.85\linewidth, width=\linewidth]{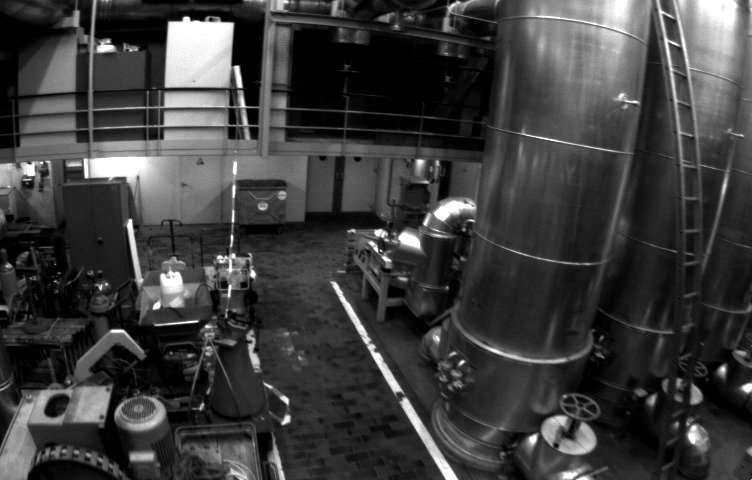}\\
		\vspace{0.05in}
		\includegraphics[height=0.85\linewidth, width=\linewidth]{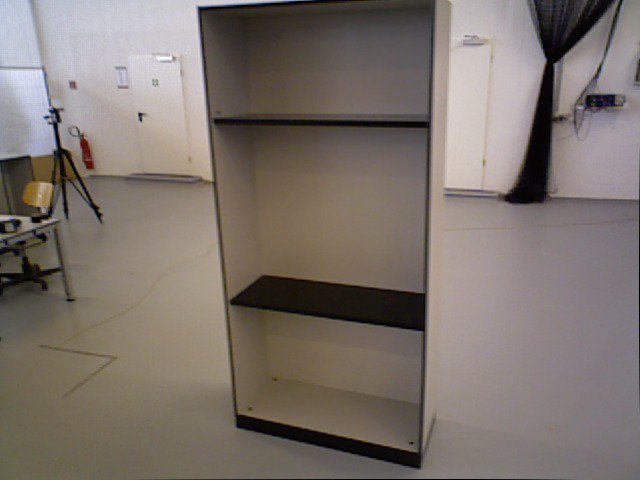}\\
		\vspace{0.05in}
		\includegraphics[height=0.85\linewidth, width=\linewidth]{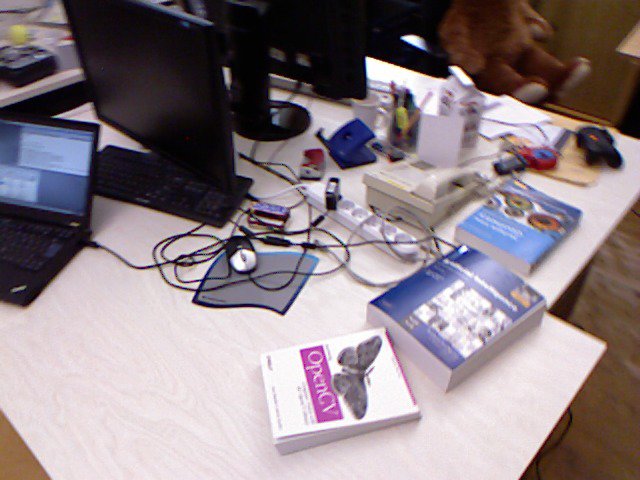}
        \end{minipage}
        \label{fig:results-a}
        }
        \subfloat[Point cloud]{
		\centering
		\begin{minipage}[t]{0.15\textwidth}
        \centering
        \includegraphics[width=\linewidth]{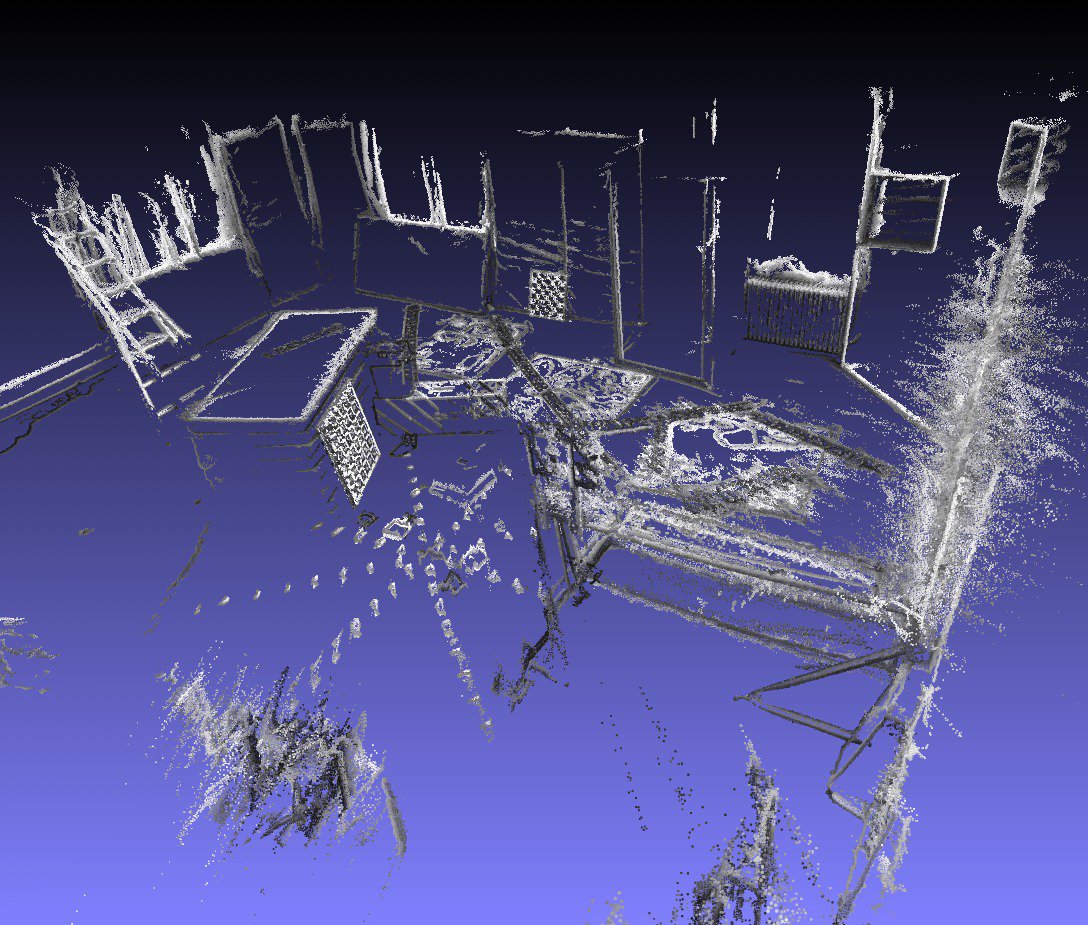}\\
        \vspace{0.05in}
		\includegraphics[width=\linewidth]{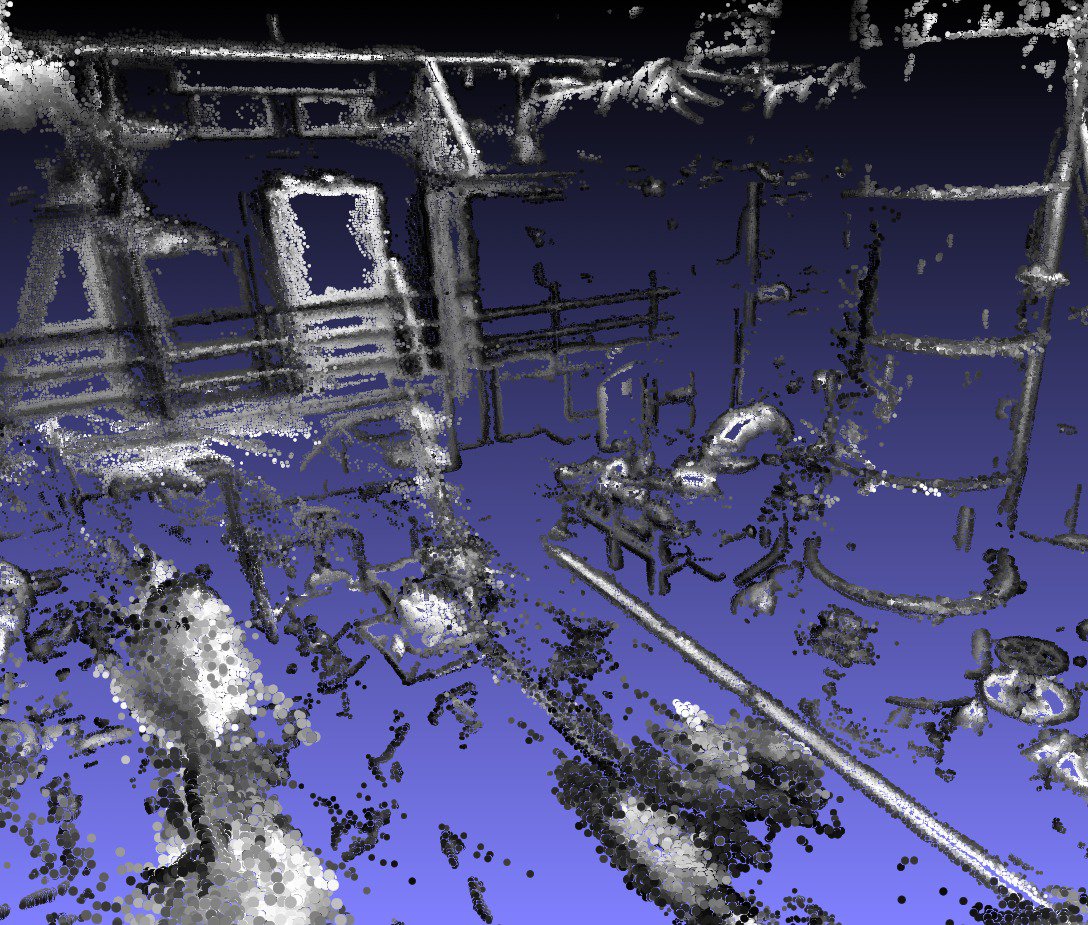}\\
		\vspace{0.05in}
		\includegraphics[width=\linewidth]{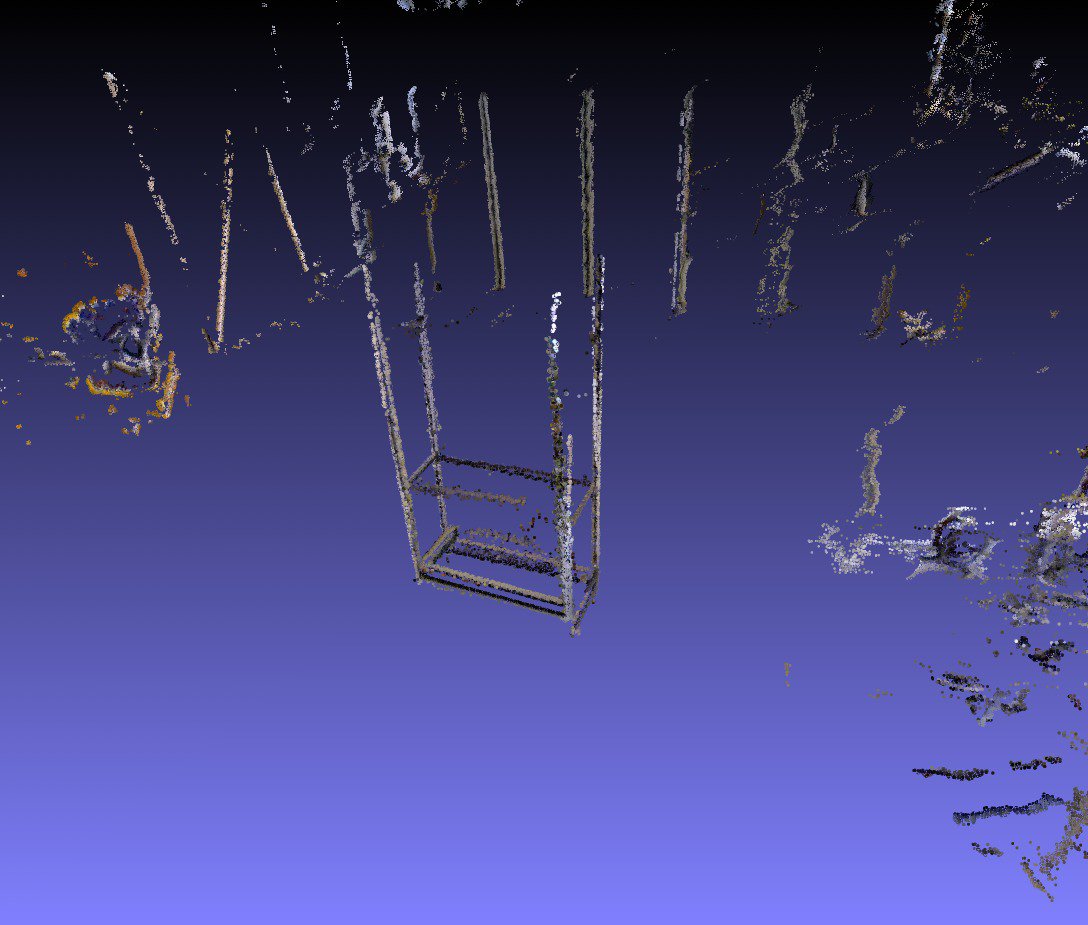}\\
		\vspace{0.05in}
		\includegraphics[width=\linewidth]{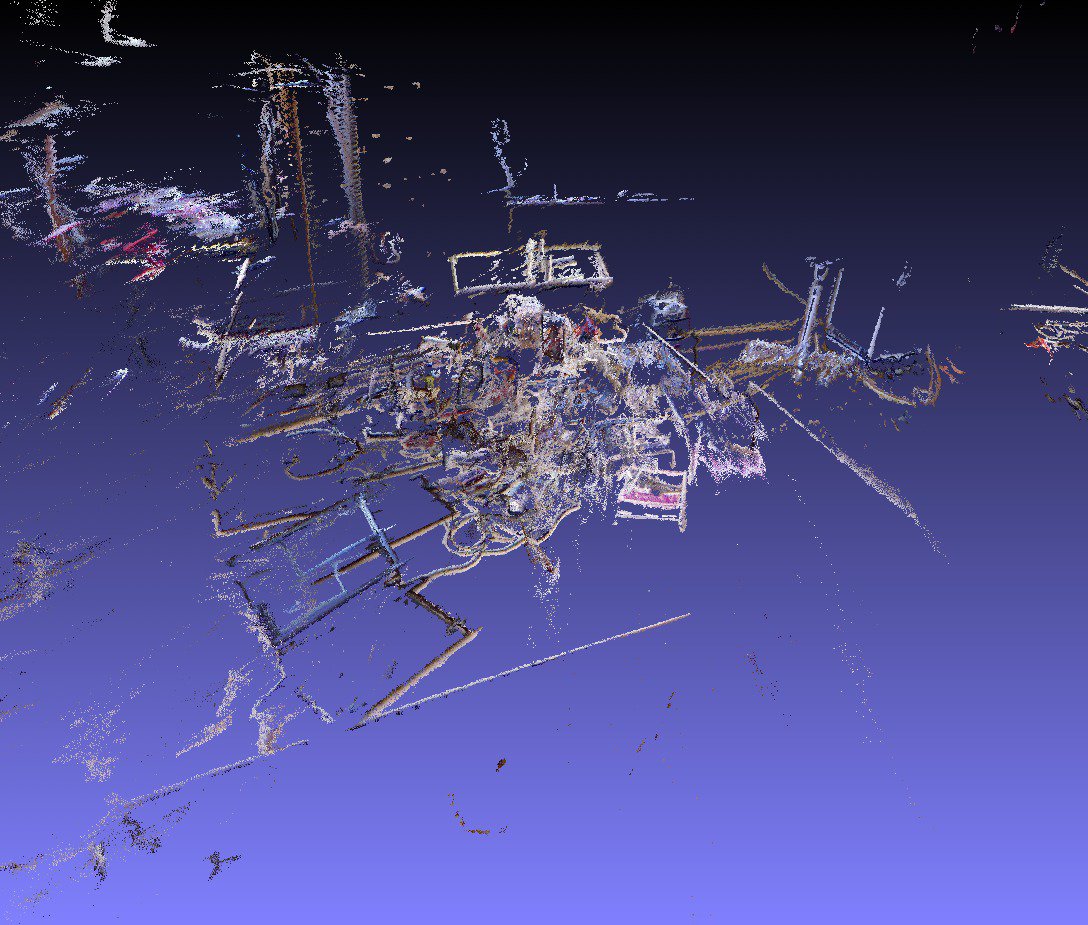}
        \end{minipage}
        \label{fig:results-b}
        }
        \subfloat[Line3d++]{
		\centering
		\begin{minipage}[t]{0.15\textwidth}
        \centering
		\includegraphics[width=\linewidth]{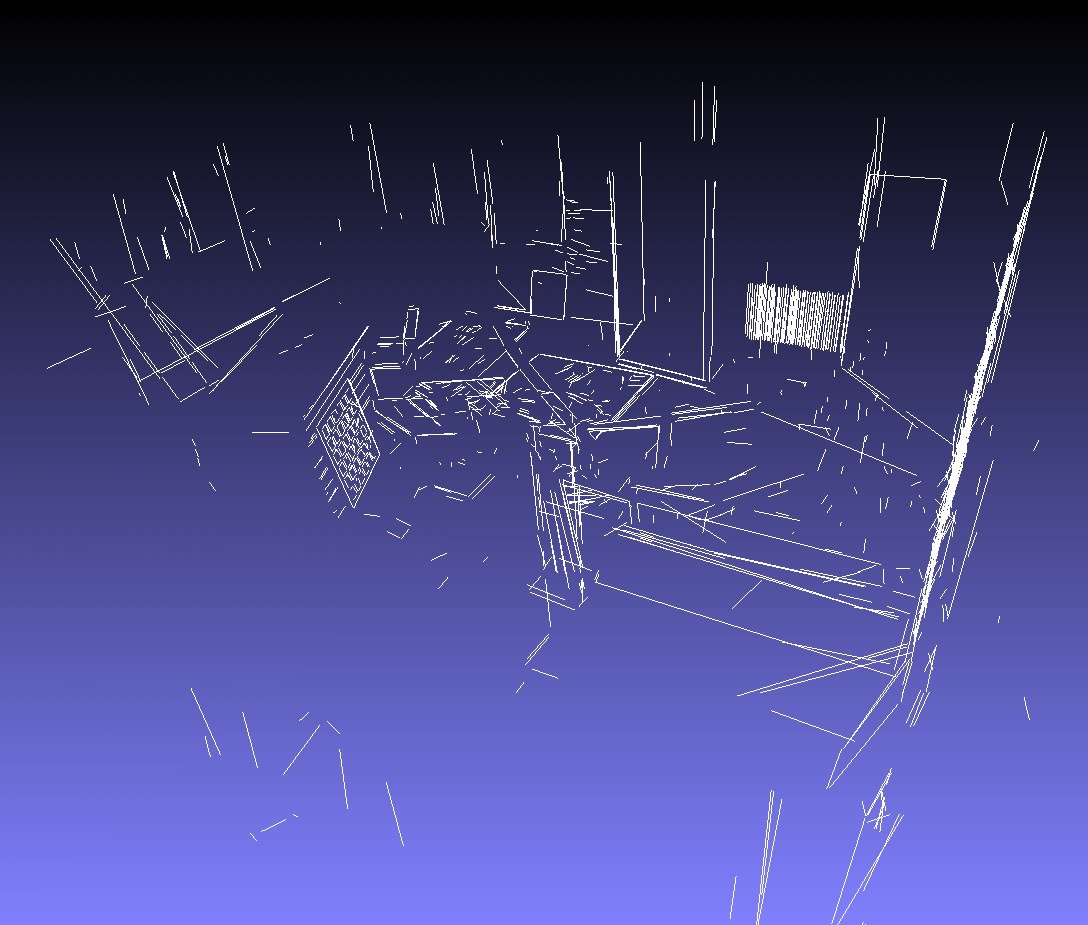}\\
		\vspace{0.05in}
		\includegraphics[width=\linewidth]{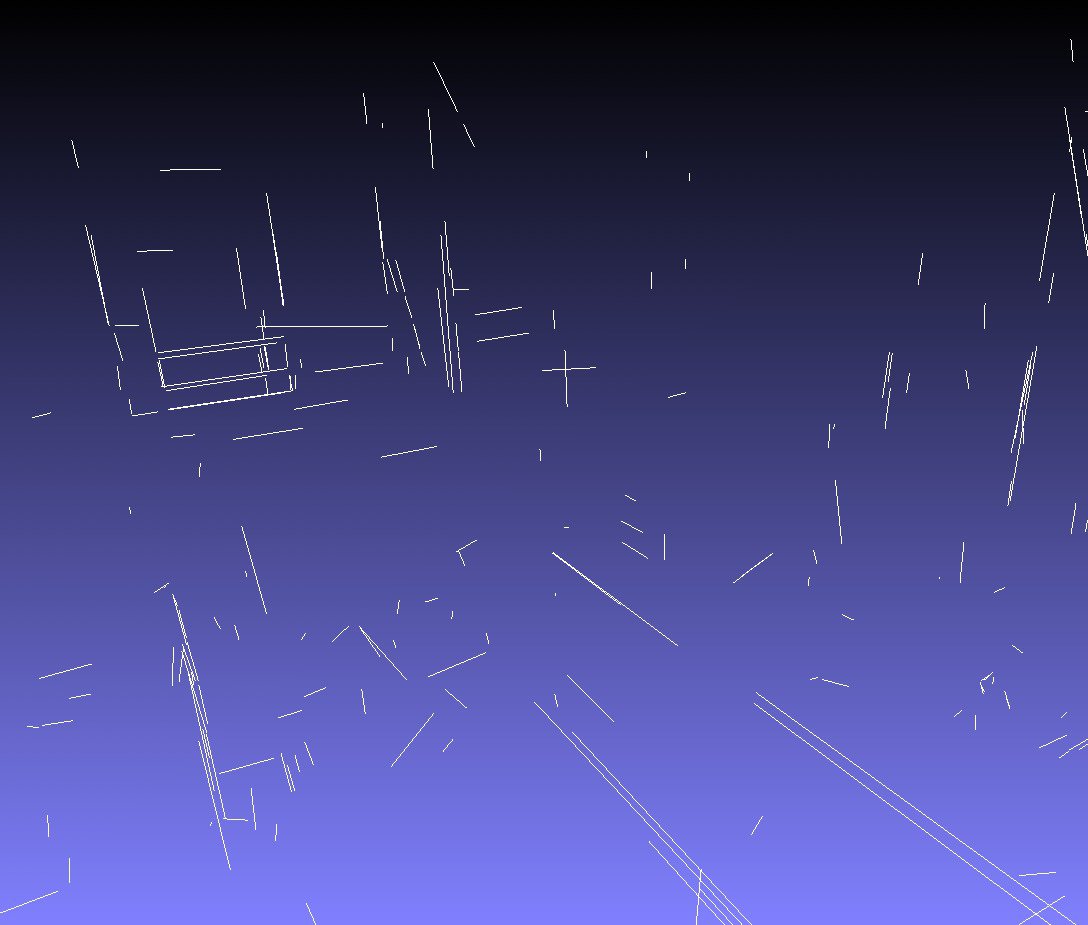}\\
		\vspace{0.05in}
		\includegraphics[width=\linewidth]{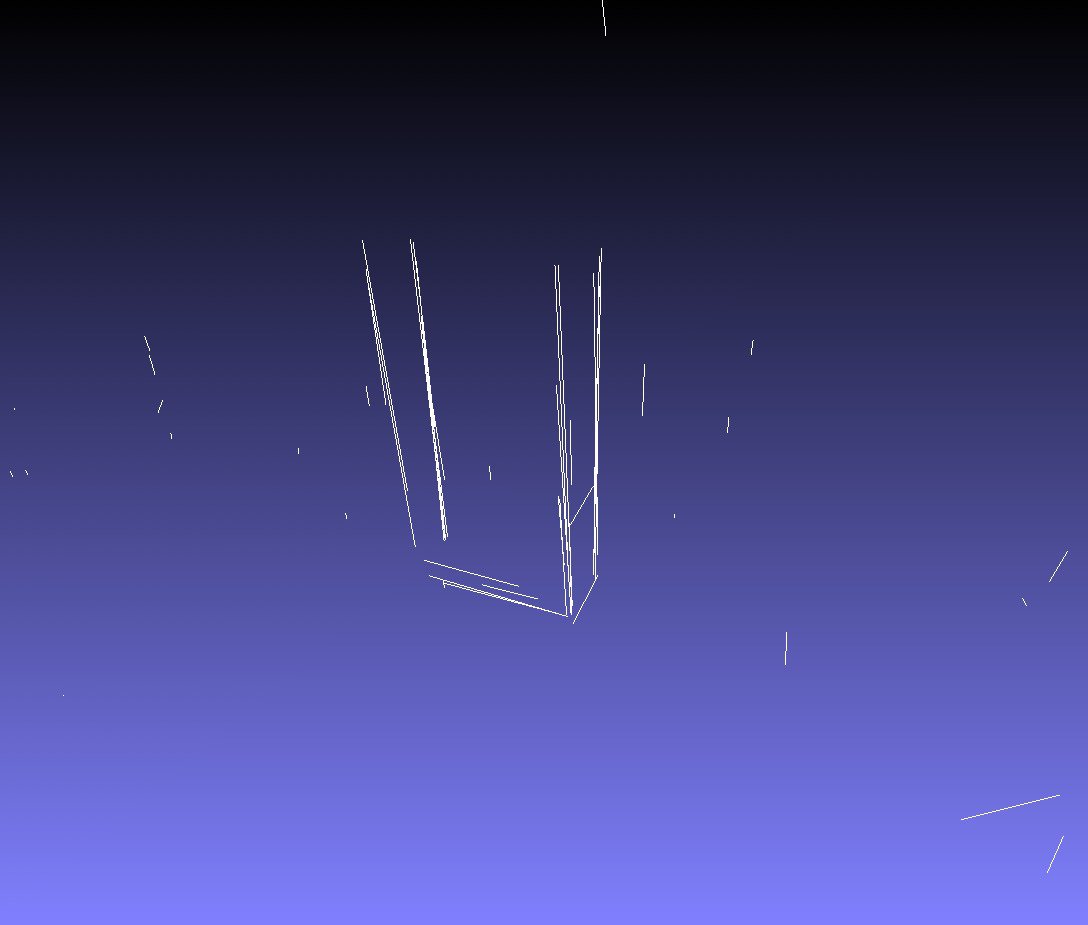}\\
		\vspace{0.05in}
		\includegraphics[width=\linewidth]{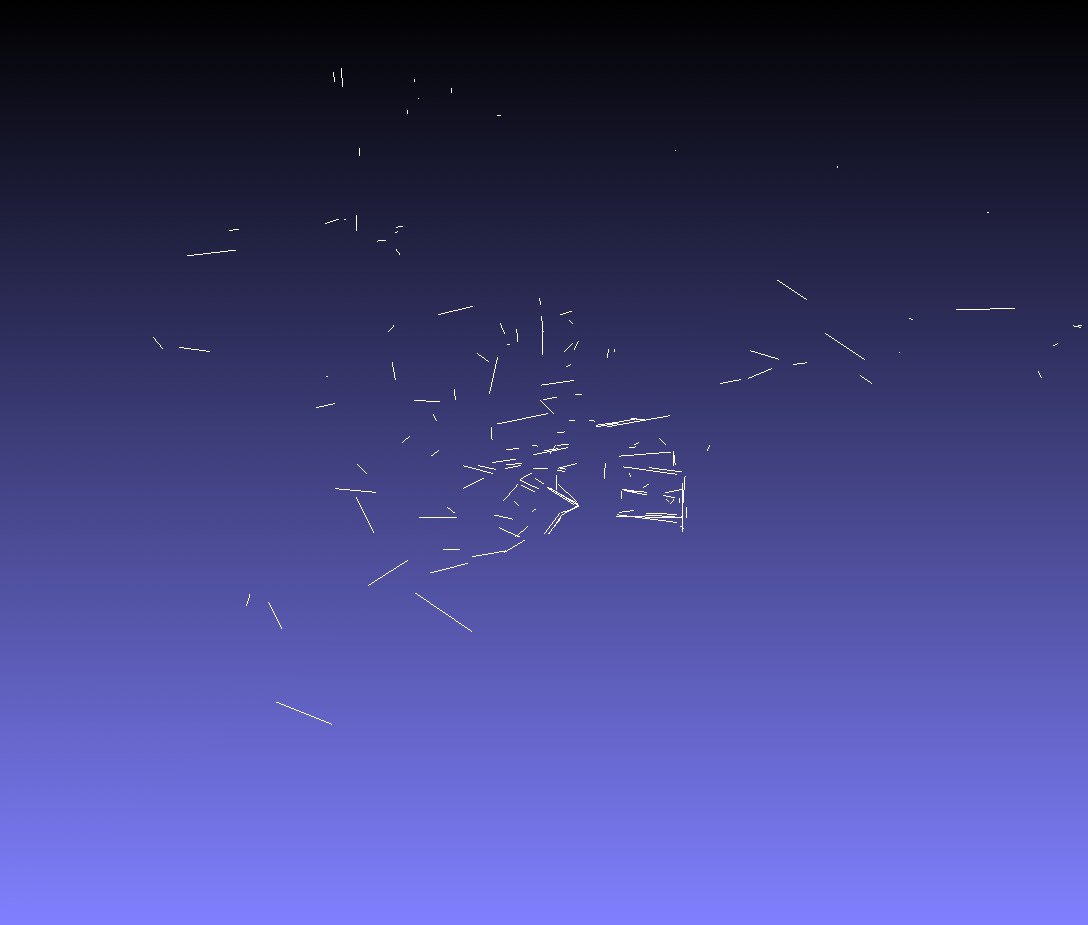}
        \end{minipage}
        \label{fig:results-c}
        }
        \subfloat[Decoupled 3D fitting]{
		\centering
		\begin{minipage}[t]{0.15\textwidth}
        \centering
		\includegraphics[width=\linewidth]{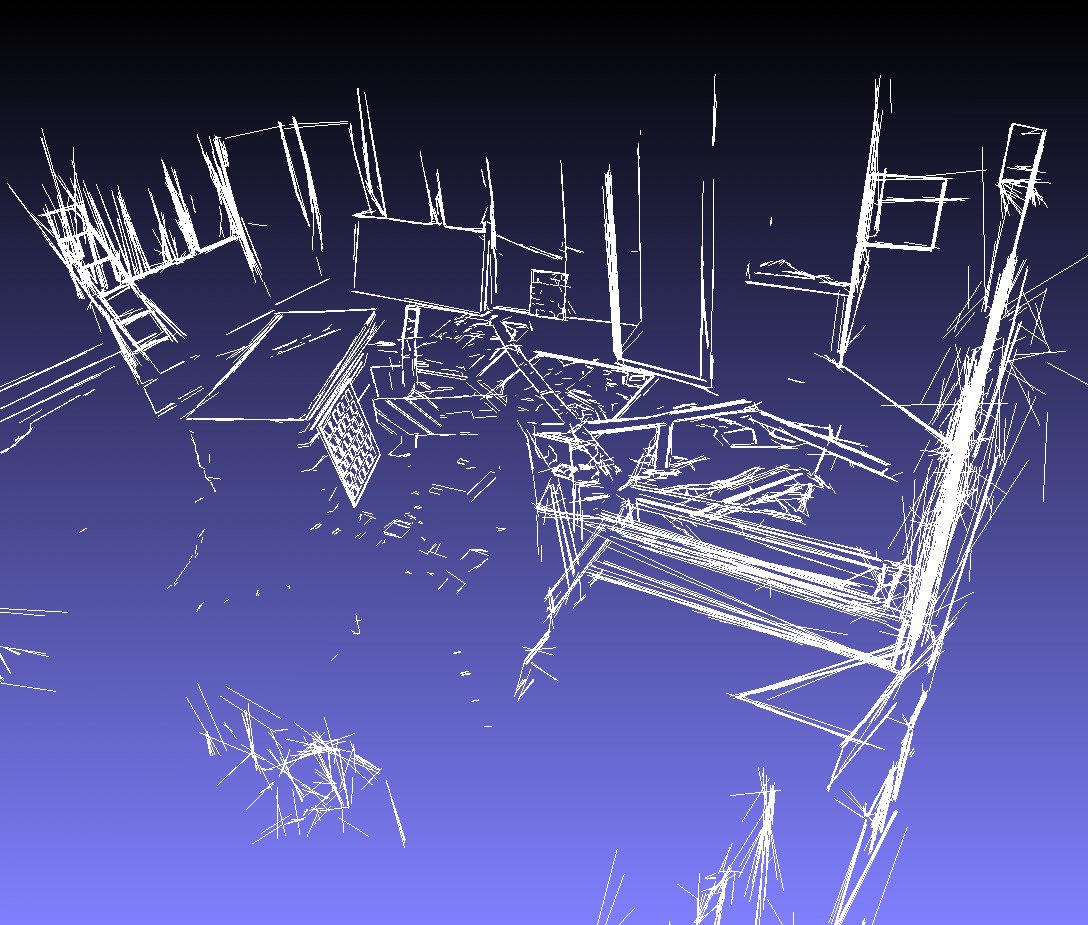}\\
		\vspace{0.05in}
		\includegraphics[width=\linewidth]{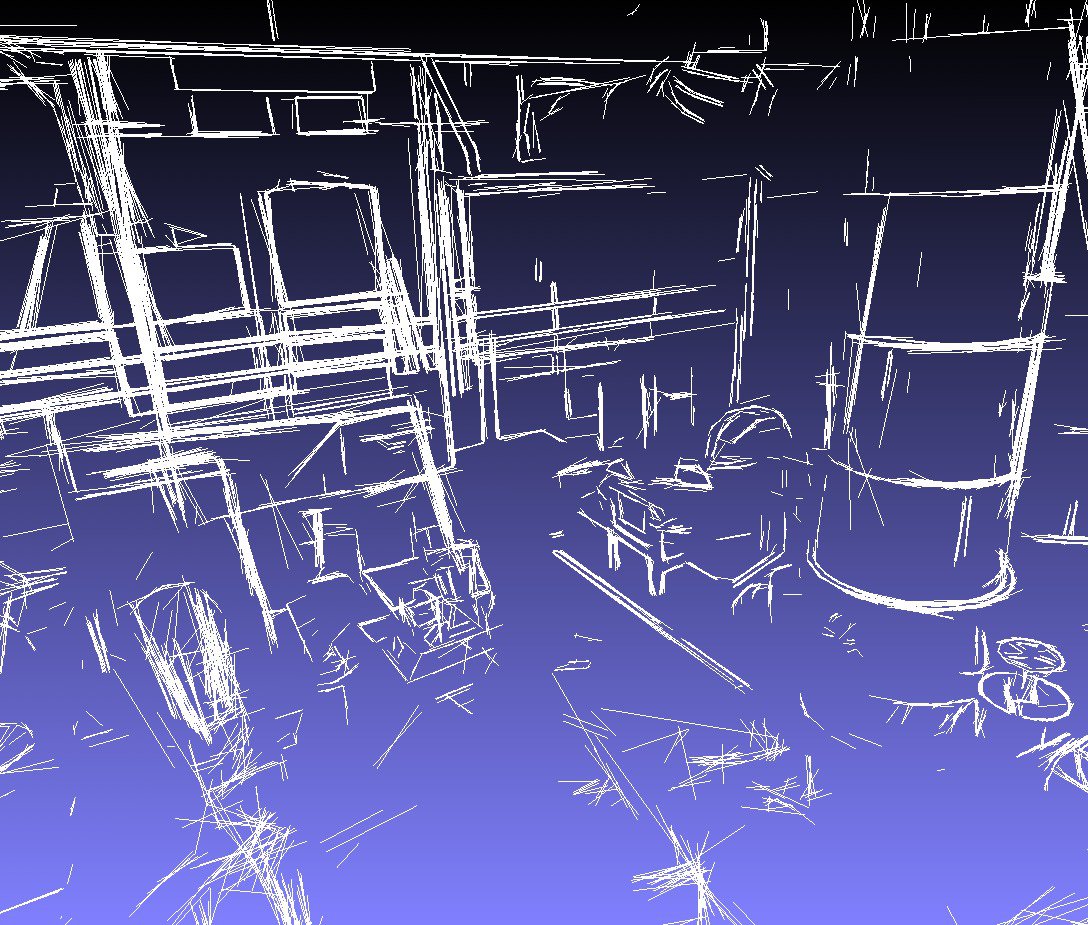}\\
		\vspace{0.05in}
		\includegraphics[width=\linewidth]{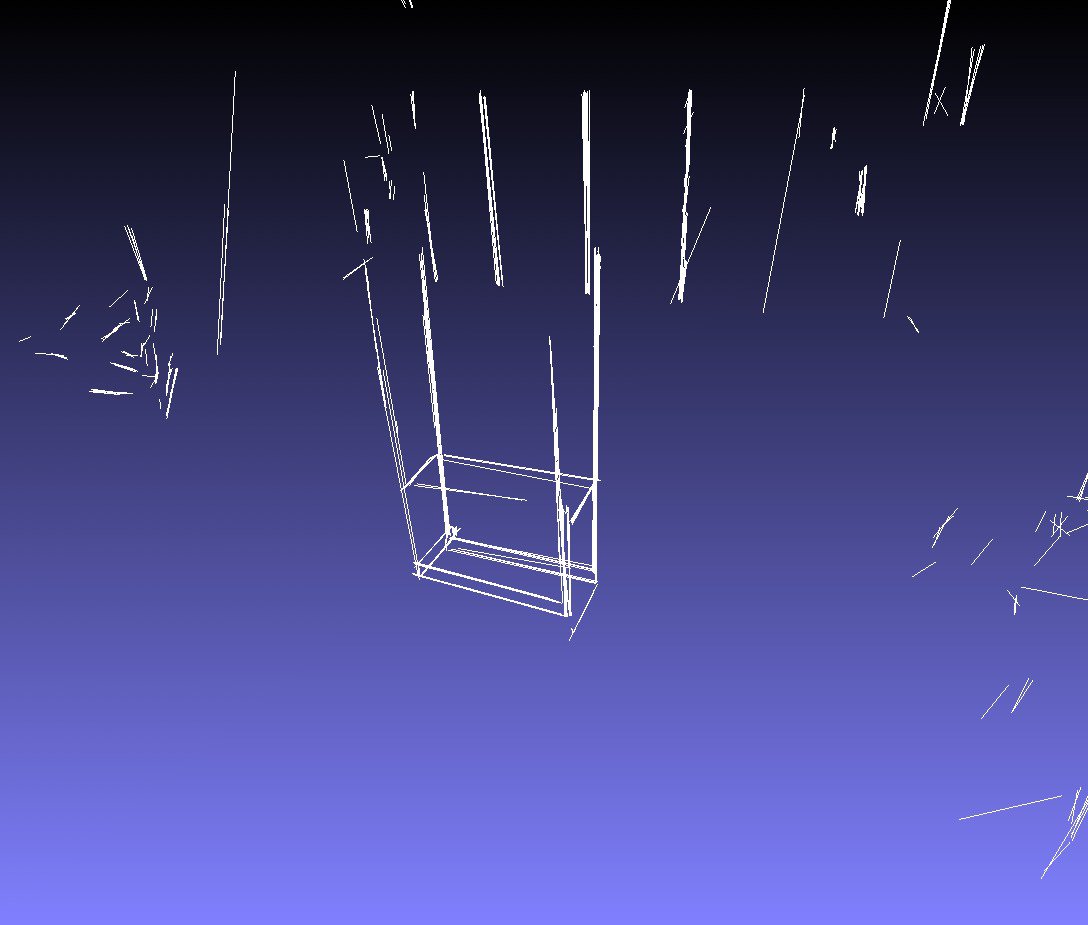}\\
		\vspace{0.05in}
		\includegraphics[width=\linewidth]{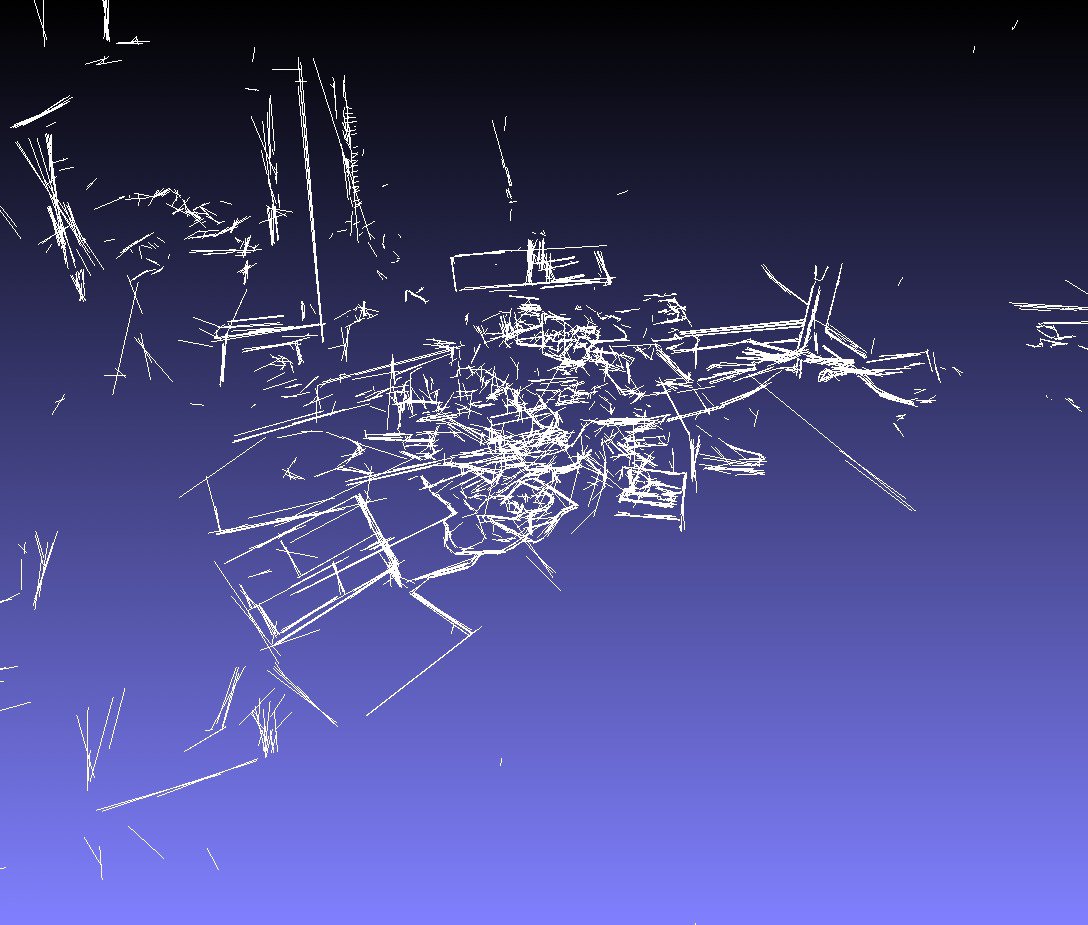}
        \end{minipage}
        \label{fig:results-d}
        }
        \subfloat[Ours w/o clustering ]{
		\centering
		\begin{minipage}[t]{0.15\textwidth}
        \centering
        \includegraphics[width=\linewidth]{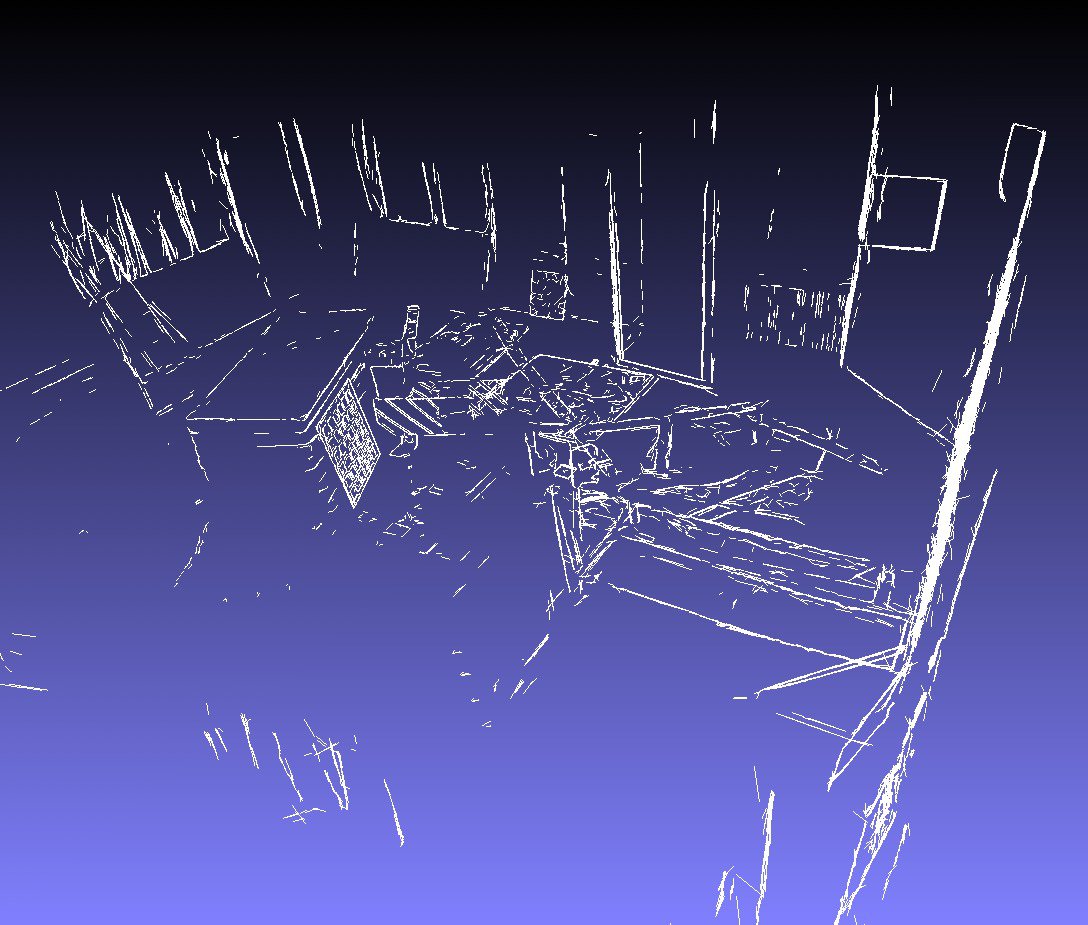}\\
        \vspace{0.05in}
		\includegraphics[width=\linewidth]{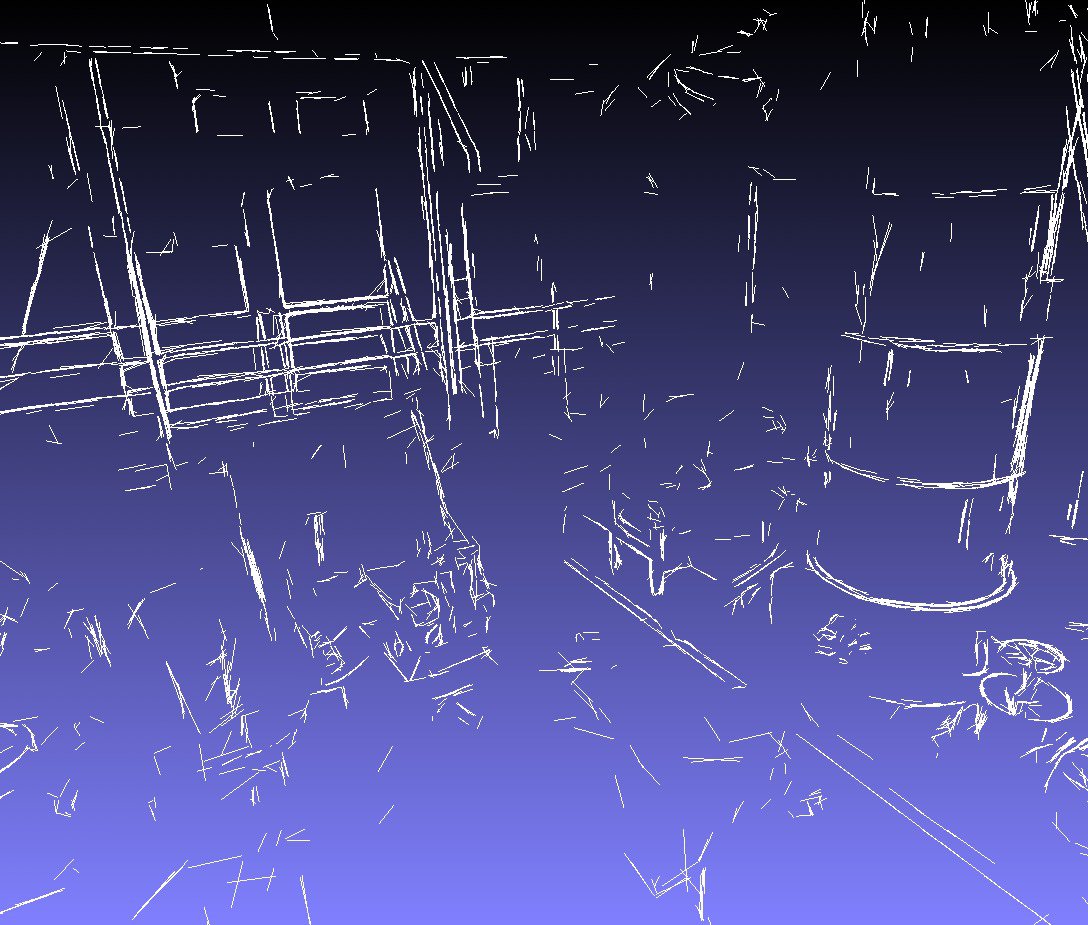}\\
		\vspace{0.05in}
		\includegraphics[width=\linewidth]{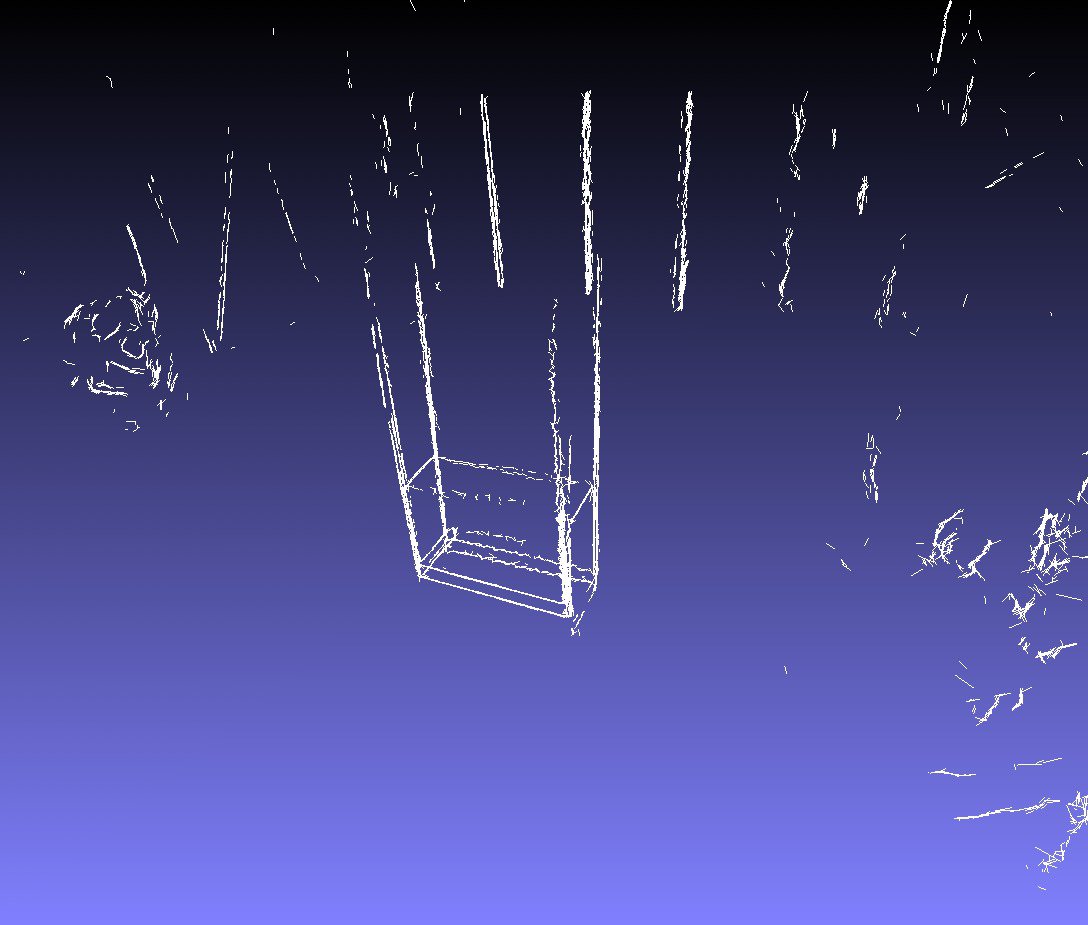}\\
		\vspace{0.05in}
		\includegraphics[width=\linewidth]{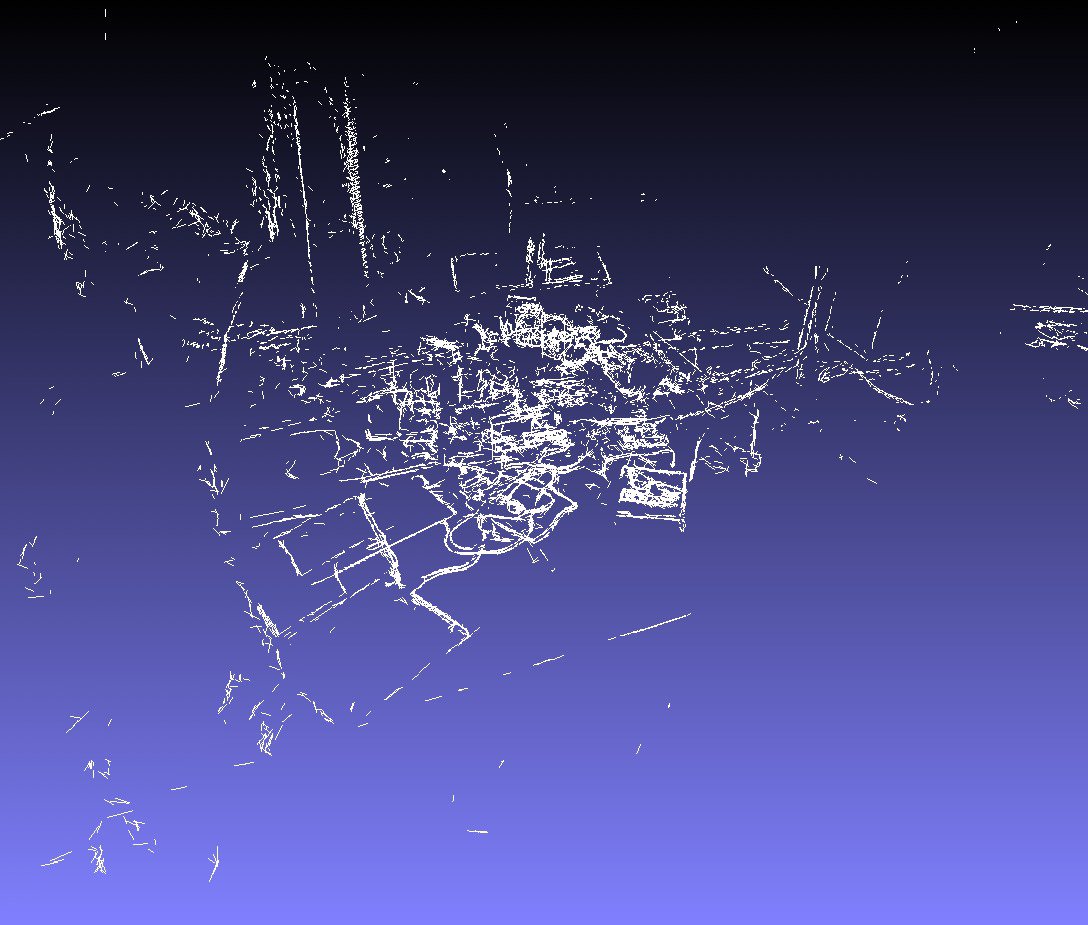}
        \end{minipage}
        \label{fig:results-e}
        }
	    \subfloat[Ours w/ clustering]{
		\centering
		\begin{minipage}[t]{0.15\textwidth}
        \centering
        \includegraphics[width=\linewidth]{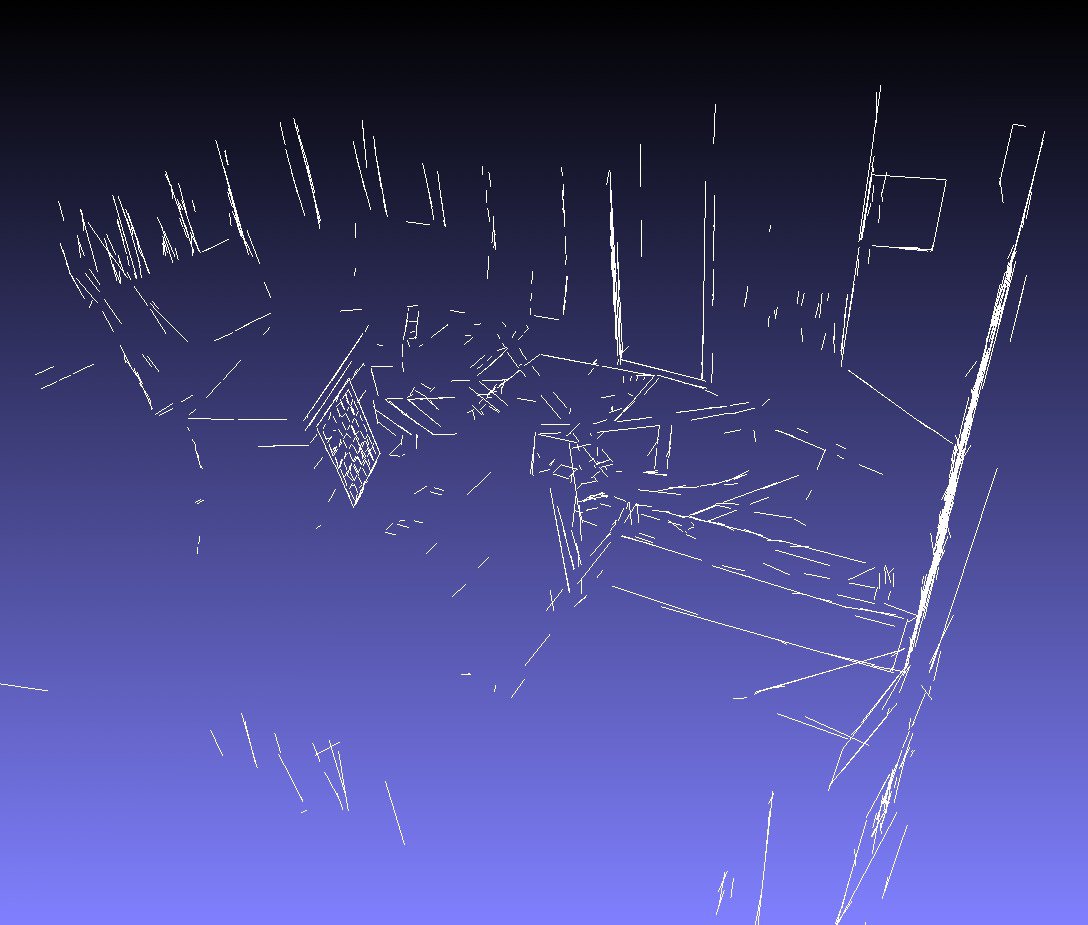}\\
        \vspace{0.05in}
		\includegraphics[width=\linewidth]{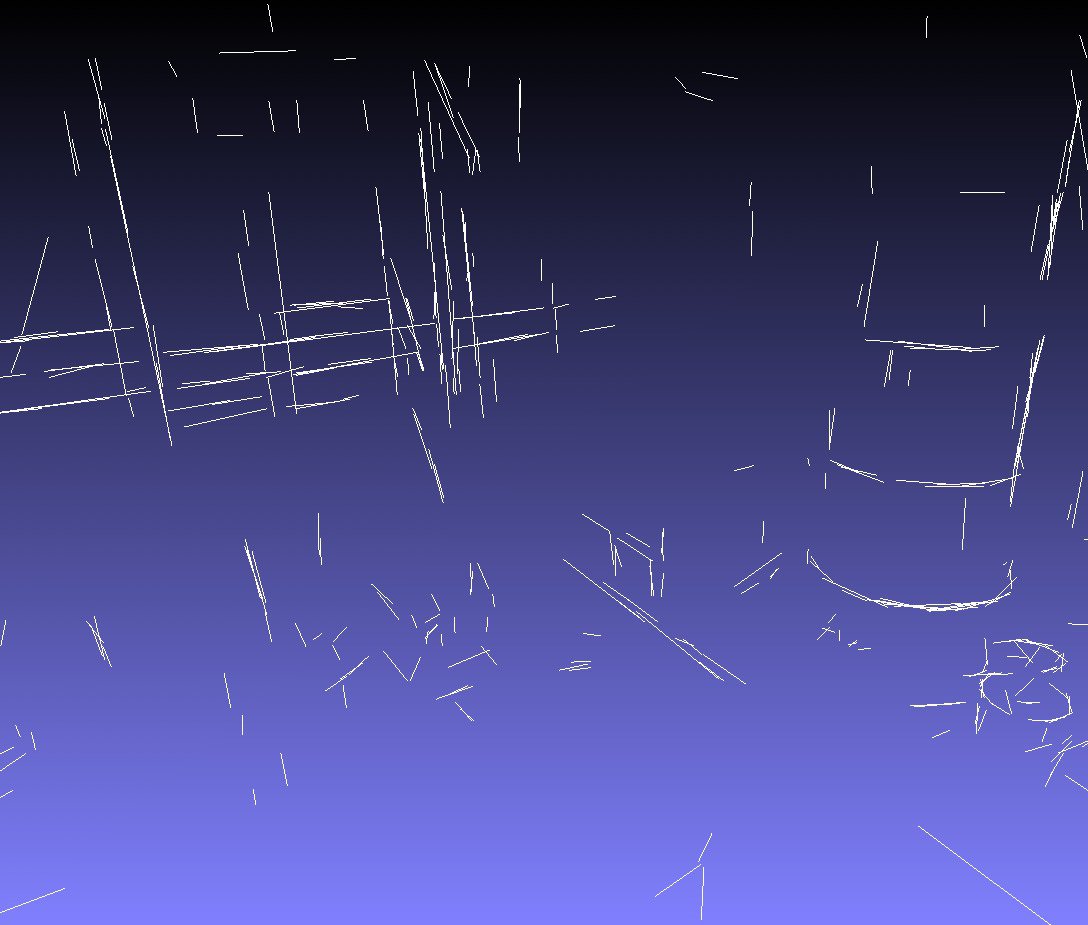}\\
		\vspace{0.05in}
		\includegraphics[width=\linewidth]{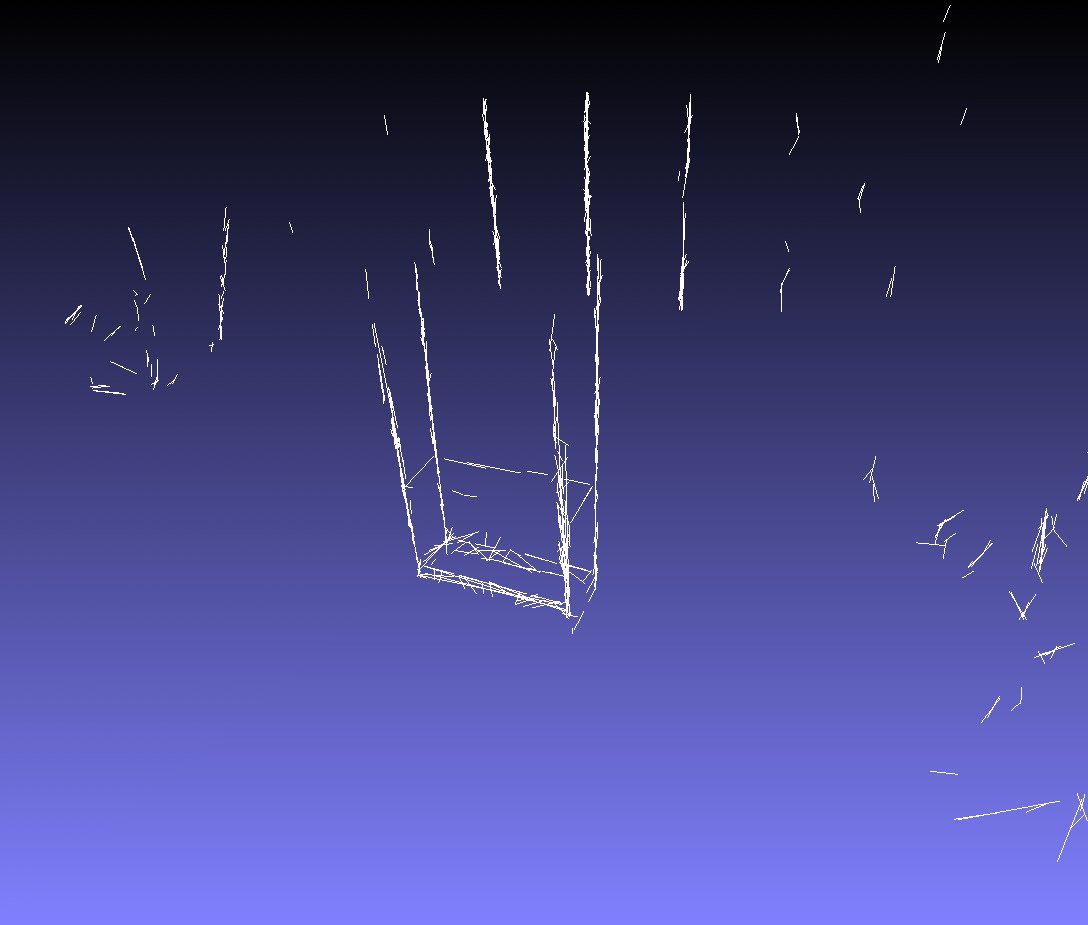}\\
		\vspace{0.05in}
        \includegraphics[width=\linewidth]{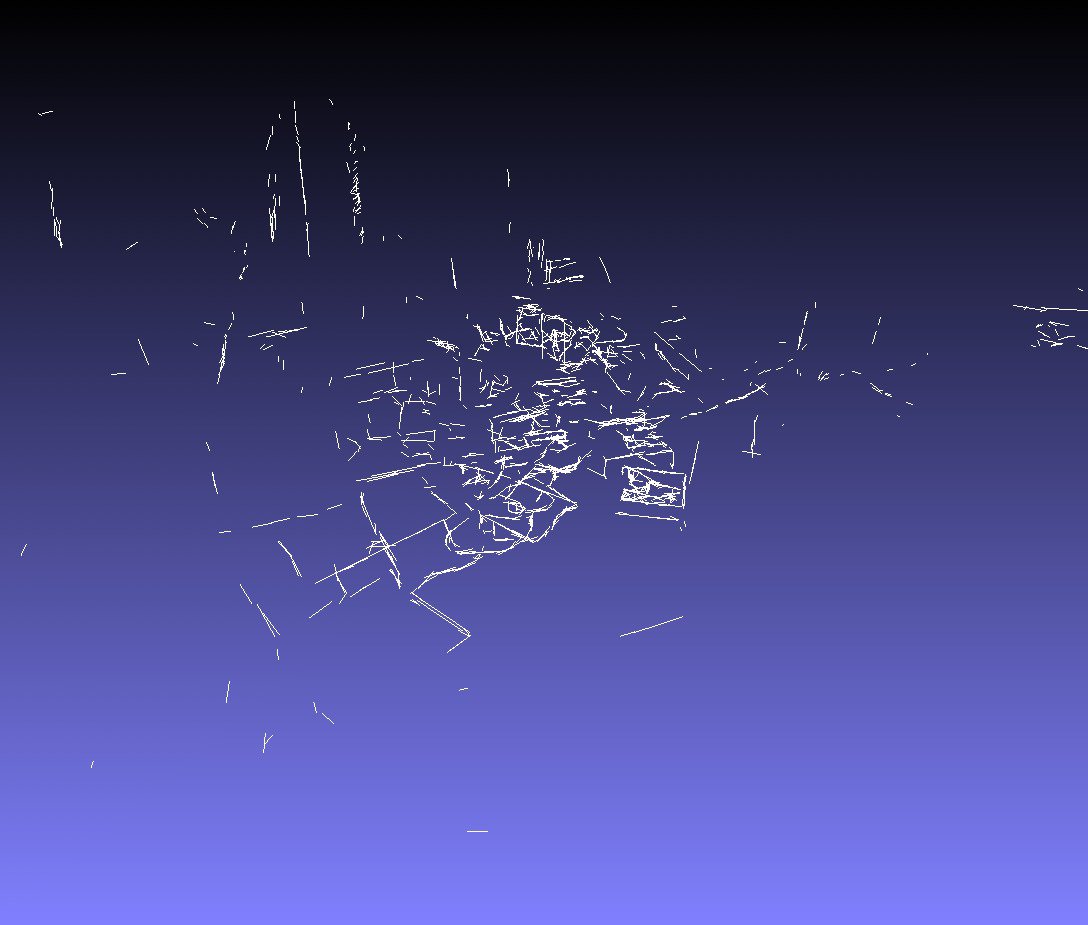}
        \end{minipage}
        \label{fig:results-f}
        }
        \caption{Experimental results. Top to Bottom (four sequences): EuRoC MAV Vicon Room 101 (\textbf{VR101}), EuRoC MAV Machine Hall 01 (\textbf{MH01}), TUM RGBD fr3-large-cabinet (\textbf{fr3-lc}), TUM RGBD fr1-room (\textbf{fr1-r}).}
		\label{fig:results}
	\end{center}
\end{figure*}


\section{Experiments}

In this section, we present the results of our 3D line segment extraction method on image sequences from the TUM RGB-D dataset \cite{DBLP:conf/iros/SturmEEBC12} and the EuRoC MAV dataset \cite{euroc}. 

\subsection{Implementation}

The experiments in this section are performed on a desktop computer with a quad-core Intel i7-6700k CPU.
We use the open source ORB-SLAM2 \cite{mur2017orb} as our base system. 
We implement the semi-dense module in C++ as described in \cite{mur2015probabilistic}.
The parameters in ORB-SLAM are kept as default, and the parameters for semi-dense module are set as presented in \cite{mur2015probabilistic}. 
Parameters in Algorithm \ref{alg:LineFitting} and incremental line segment clustering are set as follows in all our experiments: $L = 0.02*min(w,h)$, $e_1 = 0.002*min(w,h)$, $e_2 = 0.003*min(w,h)$, $\lambda_\alpha = 10$, $\lambda_d = 0.02$, $\lambda_C = 3$, where $w$ and $h$ denotes the width and height of images respectively. 

\subsection{Qualitative Comparison}


The results of our 3D line segment extraction method on some test sequences are illustrated in Figure \ref{fig:results-e} and Figure \ref{fig:results-f}.
Our results accurately fit the semi-dense point clouds shown in Figure \ref{fig:results-b}.
They still capture the major structures of the scene while reducing the number of 3D elements greatly.

\subsubsection{Line3D++} 

We first compare our results with those from Line3D++ \cite{line3D}.
The results of Line3D++ on the test sequences are shown in Figure \ref{fig:results-c}. 
In our experiments, Line3D++ uses line segments detected by EDLines \cite{akinlar2011edlines} together with the keyframe images and camera poses from ORB-SLAM to construct 3D line segments.
It utilizes geometric constraint and mutual support of line segments to match them across frames. 
In some cases, such as complex indoor environment with repetitive textures (\textit{e}.\textit{g}. sequence VR101 in Figure \ref{fig:results}),
Line3D++ can produce a large number of outliers due to the ambiguity of geometric line matching. 
On the contrary, our method is not affected by such ambiguity since it utilizes the semi-dense depth maps. 

In contrast to Line3D++, semi-dense points can cover regions with large image gradient, such as boundaries and contours, where straight lines may be absent.
Since our method takes both intensity and depth information into consideration, it is robust to outliers and it can extract shorter line segments than EDLines.
Thus, our results fit curves better and capture finer details than Line3D++, which is clearly shown in the result of sequence MH01 in Figure \ref{fig:results}. 

Moreover, our method is more robust to camera pose errors. Line3D++ can fail to correctly match line segments due to errors in the estimated camera poses, while our method is still able to reconstruct meaningful 3D line segments. It can be easily observed in the two TUM RGB-D sequences (fr3-lc and fr1-r) shown in Figure \ref{fig:results}.




\subsubsection{Decoupled 3D line segment fitting}

To further demonstrate the capability of our method, we compare it to a decoupled 3D line segment fitting method using 2D line segment given by EDLines \cite{akinlar2011edlines}.
Given detected line segments and the depth information on some of the pixels along the line segments, we can easily estimate the 3D line segment position by performing a single 2D line fitting. 
In this case, there are a fixed number of pixels on the line segment since we do not need to iteratively search along pixel chains and extend line segments.
Therefore, we can efficiently perform RANSAC in 2D to remove outliers before the line fitting process.  
With the fitted line, we compute the 3D location of the endpoints and reconstruct the 3D line segment.
Note the result of this method is equivalent to directly performing a RANSAC in 3D to fit all 3D points on the line segment.
However, fitting a line in 2D is faster because fewer parameters are required to represent the line and the search space is much smaller.

\begin{figure}
	\begin{center}
	
        \subfloat[Decoupled fitting using EDLines]{
        \centering
		\begin{tikzpicture}[zoomboxarray, zoomboxarray columns=1, zoomboxarray rows=1]
            \node [image node] { \includegraphics[width=0.108\textwidth]{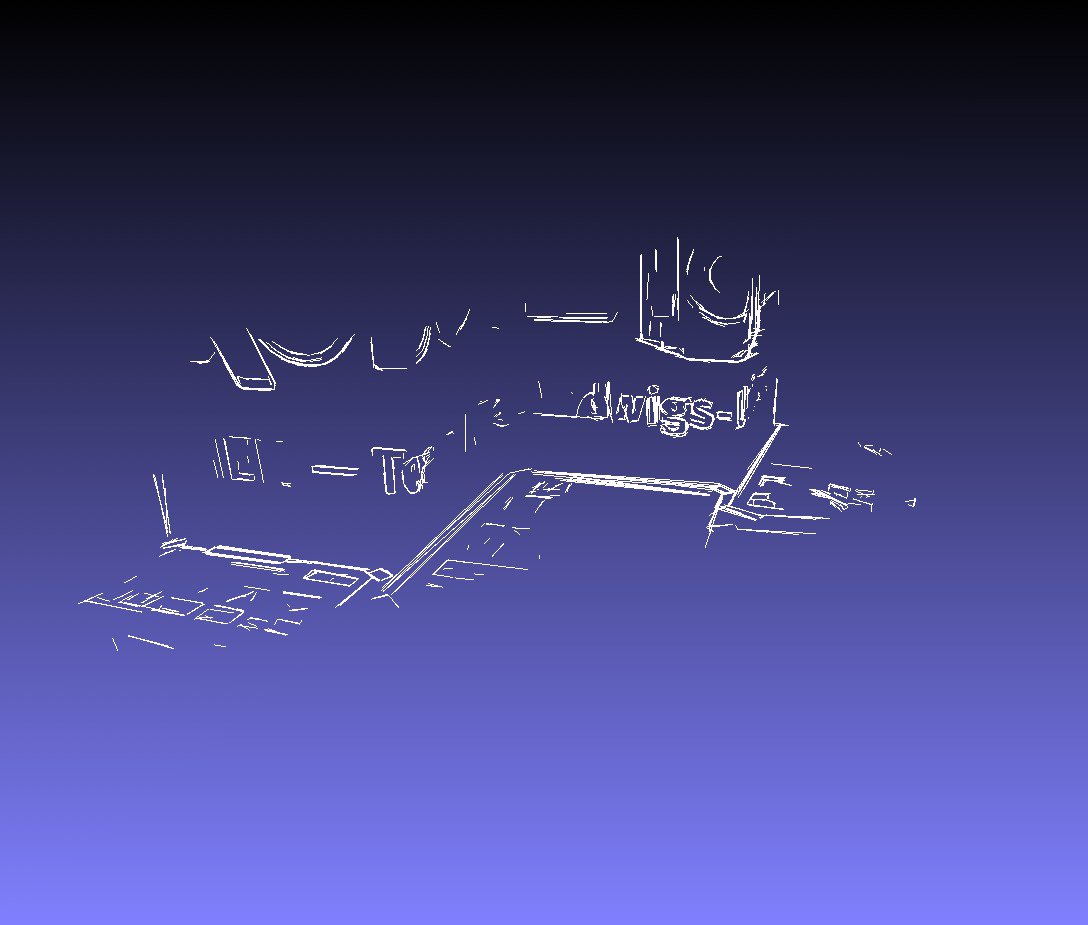} };
            \zoombox[magnification=8]{0.4,0.4}
        \end{tikzpicture}
        }
        ~
        \subfloat[Our edge-aided fitting]{
        \centering
		\begin{tikzpicture}[zoomboxarray, zoomboxarray columns=1, zoomboxarray rows=1]
            \node [image node] { \includegraphics[width=0.108\textwidth]{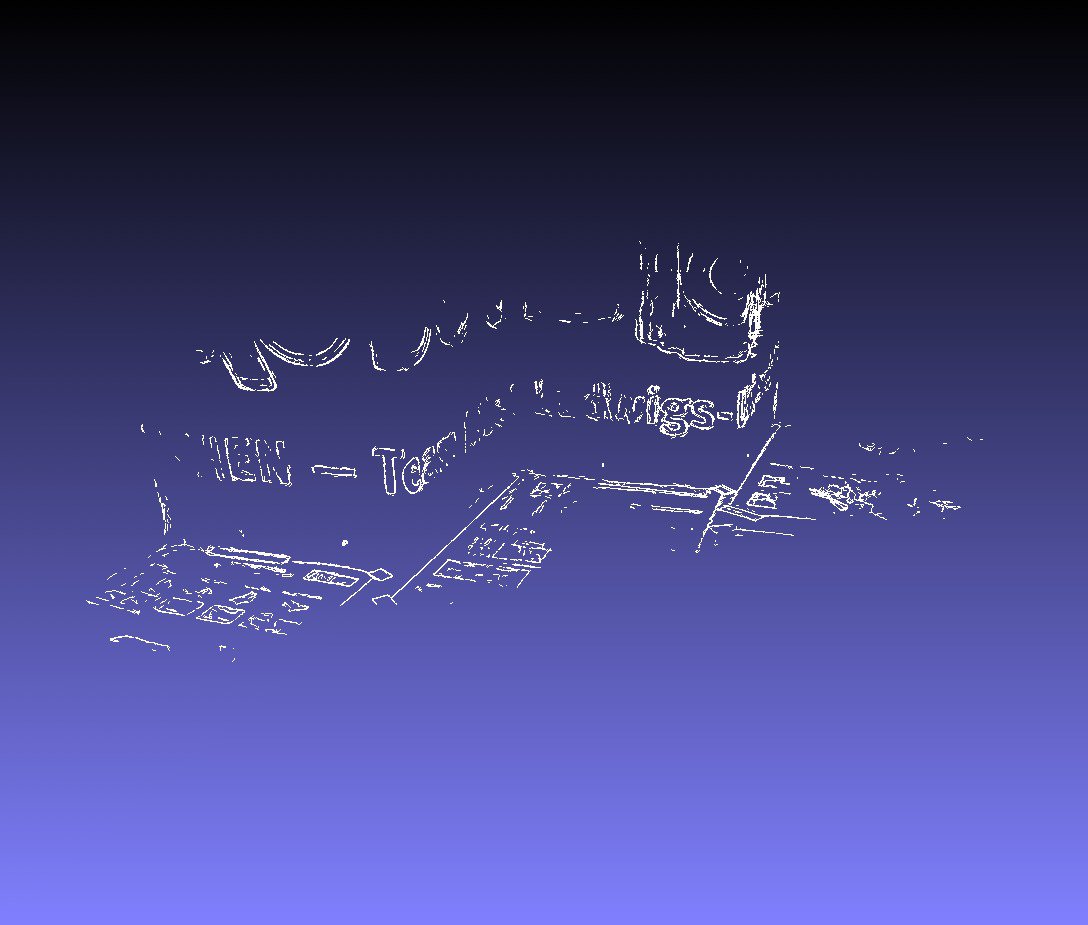} };
            \zoombox[magnification=8]{0.4,0.4}
        \end{tikzpicture}
        }
        
		\caption{Comparison of results of sequence TUM RGB-D fr3-structure-texture-near.}
		\label{fig:ShortLongLine}
	\end{center}
\end{figure}

The results of decoupled line segment fitting are presented in Figure \ref{fig:results-d}.
Compared to the edge aided 3D line fitting which tries to utilize pixel position and depth simultaneously, the decoupled fitting essentially fits lines in the image plane and depth plane in two steps.
The error from line fitting in the image plane will be propagated to the error of 3D line segment position, which result in an inaccurate reconstruction compared to our method, as shown in Figure \ref{fig:ShortLongLine}. 
It is worth mentioning that the decoupled fitting tends to generate much longer segments since only the pixel position is considered in the image plane line fitting process. 
Longer segments will make the propagated error even worse because the total error of line segments in image space might be larger.
Another source of error is that EDLines may detect long line segments which are not continuous lines in 3D space. 
Trying to fit a single 3D line segment onto the 2D segment in this case will result in a large error. 
On the other hand, in our method, if either of the two errors of line fitting grows higher than the threshold, we will stop the line fitting and start a new line fitting process.
In this way, the errors accumulated from image plane and depth plane are bounded, and therefore prevent the line segments from being far away from the 3D points.


\subsection{Quantitative Comparison}

\begin{table}[bp]
\caption{Average distance of vertices to ground truth surface}
\label{tb:distance}
\begin{center}
\begin{tabular}{|l|c|c|c|c|}
\hline
 Method & Vicon Room 101 & Vicon Room 201 \\
\hline\hline
 Line3D++                               & 84.10 mm    & 78.63 mm    \\
 Decoupled 3D fitting                   & 21.48 mm    & 23.45 mm  \\
 Ours w/o clustering              & 13.91 mm    & 17.41 mm \\
 Ours w/ clustering               & 13.93 mm    & 17.63 mm   \\
 \hline
\end{tabular}
\end{center}
\end{table}

\subsubsection{Distance to surface}

To demonstrate the accuracy of our method, we compute the average distance of line segment endpoints to the ground truth surface in two EuRoC MAV sequences, as shown in Table \ref{tb:distance}.
We take the provided precise 3D scanning of environment as ground truth.
Since the output of ORB-SLAM have coordinates different from the ground truth surface data, we estimate the global Euclidean transform and scale change by performing ICP to align the semi-dense point cloud to the ground truth point cloud.
The same Euclidean transform and scale change are applied to all the output line segments, so that all the distances are calculated in the coordinates of the ground truth data.
It can be seen in Table \ref{tb:distance} that the results of our method fit to the surfaces better than other methods.

\subsubsection{Compactness} 

For easy handling and manipulation, it is desired to have fewer 3D elements while they can still represent most of the environment.
In the surface reconstruction pipeline, a smaller number of vertices will also greatly reduce the running time.
As shown in Table \ref{tb:vertex}, the point clouds are greatly simplified with our edge aided 3D line segment fitting algorithm.
The results are simplified further
using our 3D line segments clustering process.
Note that although Line3D++ produces the fewest number of vertices in the reconstruction, the completeness of reconstruction is generally worse than our method as shown in Figure \ref{fig:results}.


\begin{table}[bp]
\caption{Number of vertices in the reconstruction}
\label{tb:vertex}
\begin{center}
\begin{tabular}{|l|c|c|c|c|}
\hline
 Data & VR101 & MH01 & fr3-lc & fr1-r \\
\hline\hline
 Point cloud                 & 2361598  & 3252467  & 263637  & 1044752\\
 Line3D++                               & 2832    & 2354     & 124     & 330 \\
 Decoupled 3D fitting                   & 15994    & 42416    & 1106    & 9966\\
 Ours w/o clustering              & 34958   & 41718   & 15760   & 42624 \\
 Ours w/ clustering               & 2396    & 2810    & 1304    & 3334\\
 \hline
\end{tabular}

\end{center}
\end{table}

\subsubsection{Running Time}
The average running time of our edge-aided 3D line segments fitting on the sequences shown in Figure \ref{fig:results} is 7.40ms per keyframe, while the decoupled 3D fitting takes 10.42ms per keyframe. 
Our line segment fitting method is run-time efficient while utilizing large amount of depth information.
Compared to the running time of edge aided 3D fitting, decoupled 3D fitting requires additional computation time for performing RANSAC.
Because the segments are usually much longer in decoupled 3D line segments fitting, RANSAC is necessary in order to obtain a good fit for the larger pixel sets on line segments.
Our 3D line segment fitting algorithm is linear in the number of line segments and is fast enough to be real-time. However, our clustering process is relatively slower.
The complexity of clustering a single line segment is $O(M)$, where $M$ is the number of existing clusters. In the worst case, $M$ can be the same as the number of line segments $N$. Therefore, the complexity of the clustering process in a sequence is $O(N^2)$.


\begin{figure}
	\begin{center}
		\subfloat[Sample images]{
		\centering
		\begin{minipage}[t]{0.48\textwidth}
        \centering
		\includegraphics[width=0.45\textwidth]{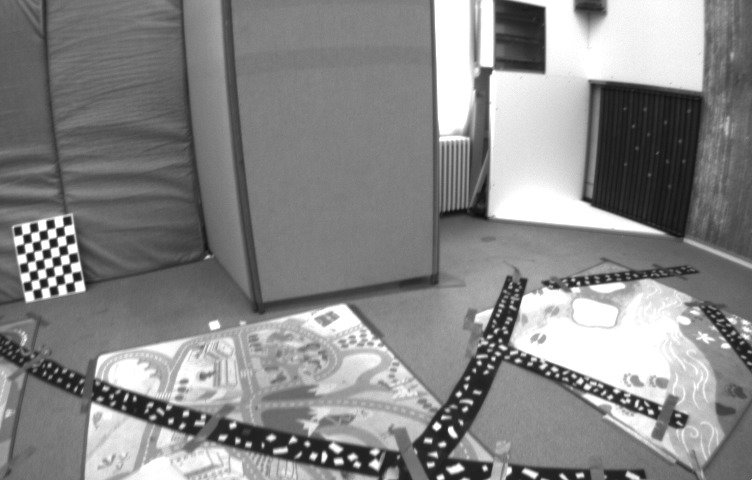}
		~
		\includegraphics[width=0.45\textwidth]{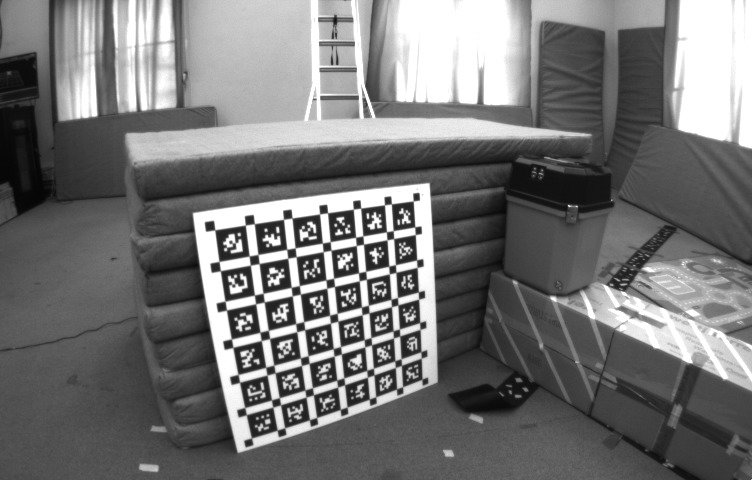}
        \end{minipage}
        }\\
        \subfloat[Surface with map points in ORB-SLAM]{
        \centering
        \begin{minipage}[t]{0.48\textwidth}
        \centering
		\includegraphics[width=0.45\textwidth]{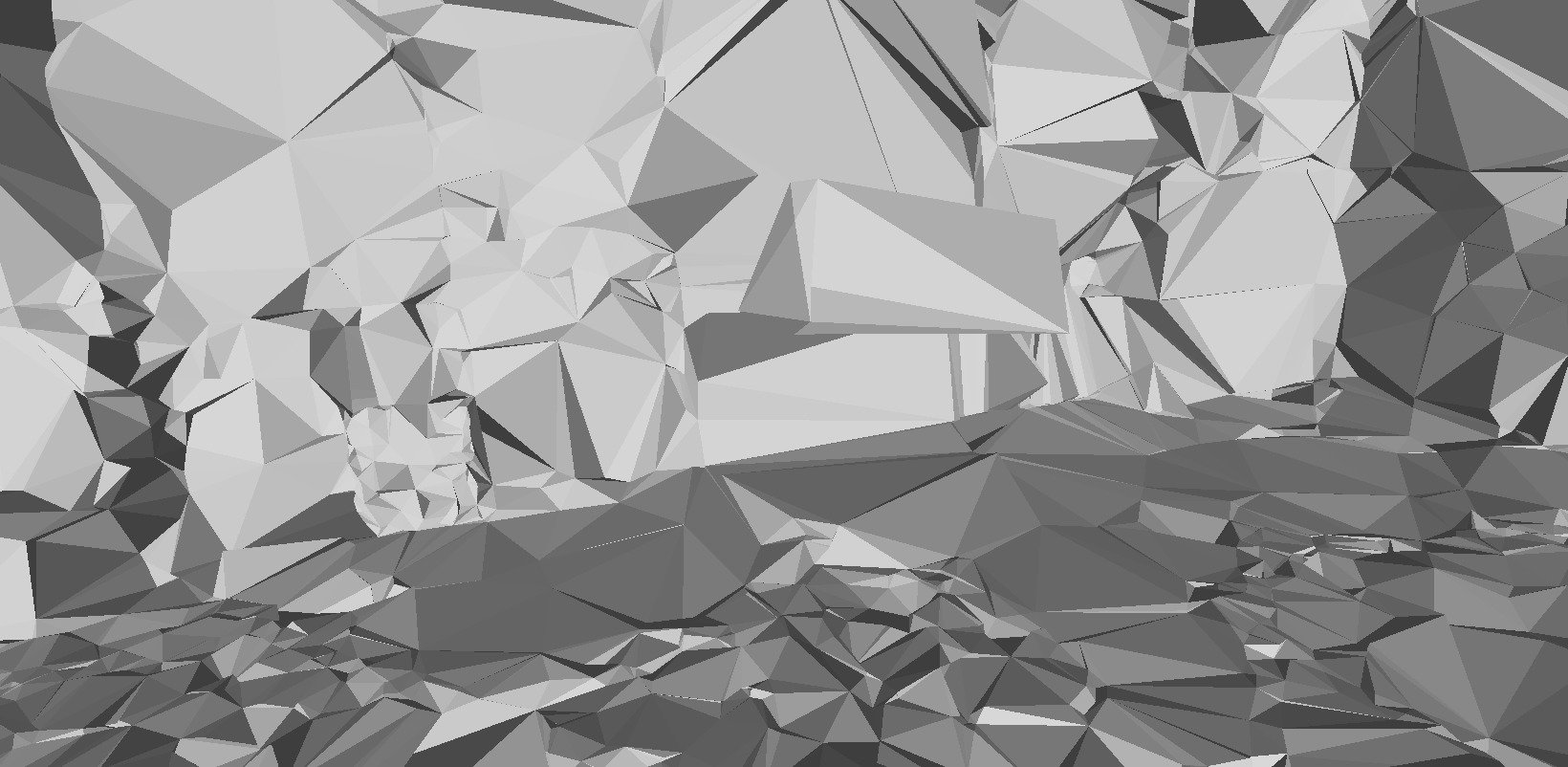}
		~
		\includegraphics[width=0.45\textwidth]{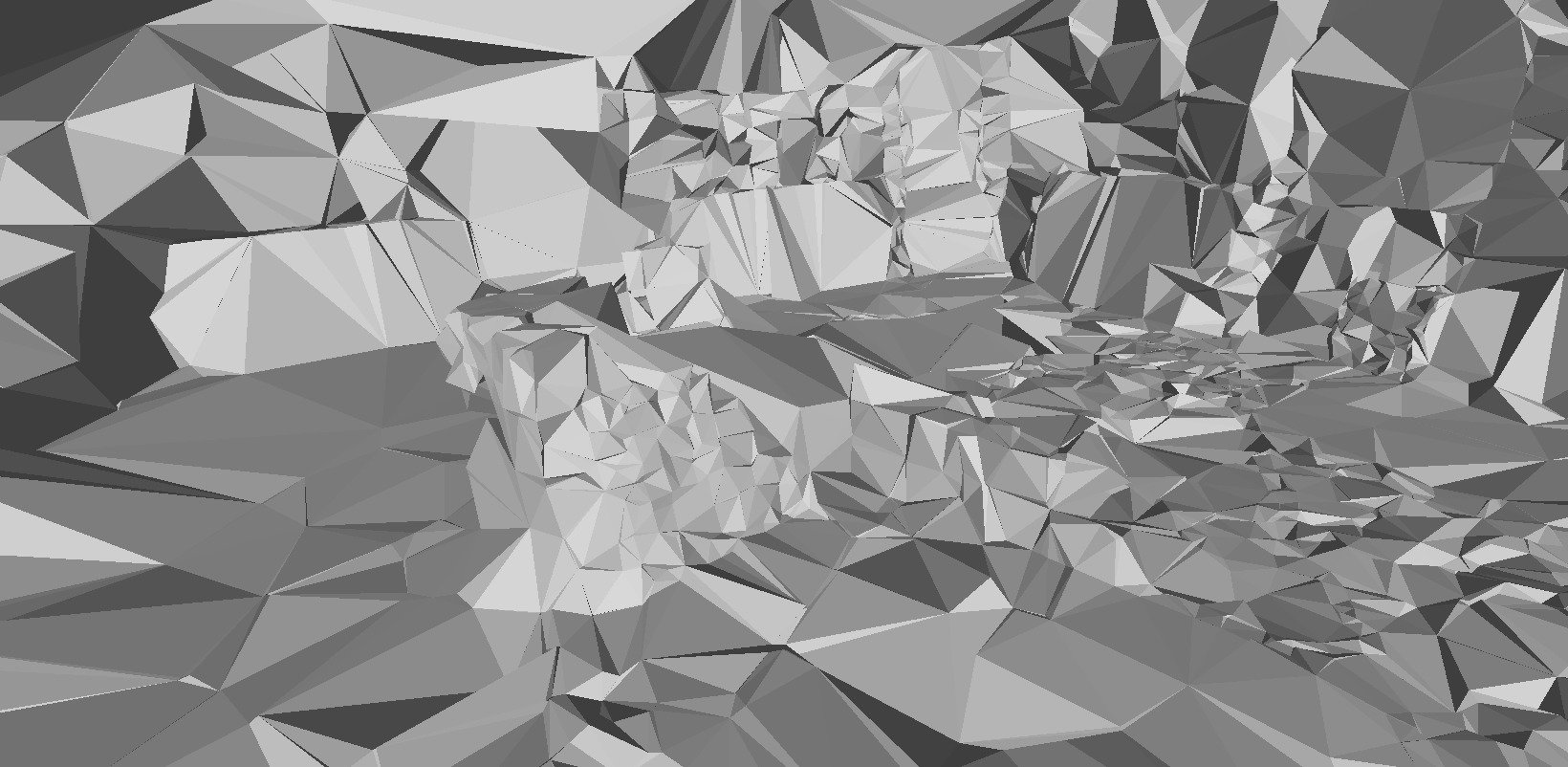}
        \end{minipage}
        }\\
        \subfloat[Surface with our line segments endpoints]{
        \centering
        \begin{minipage}[t]{0.48\textwidth}
        \centering
		\includegraphics[width=0.45\textwidth]{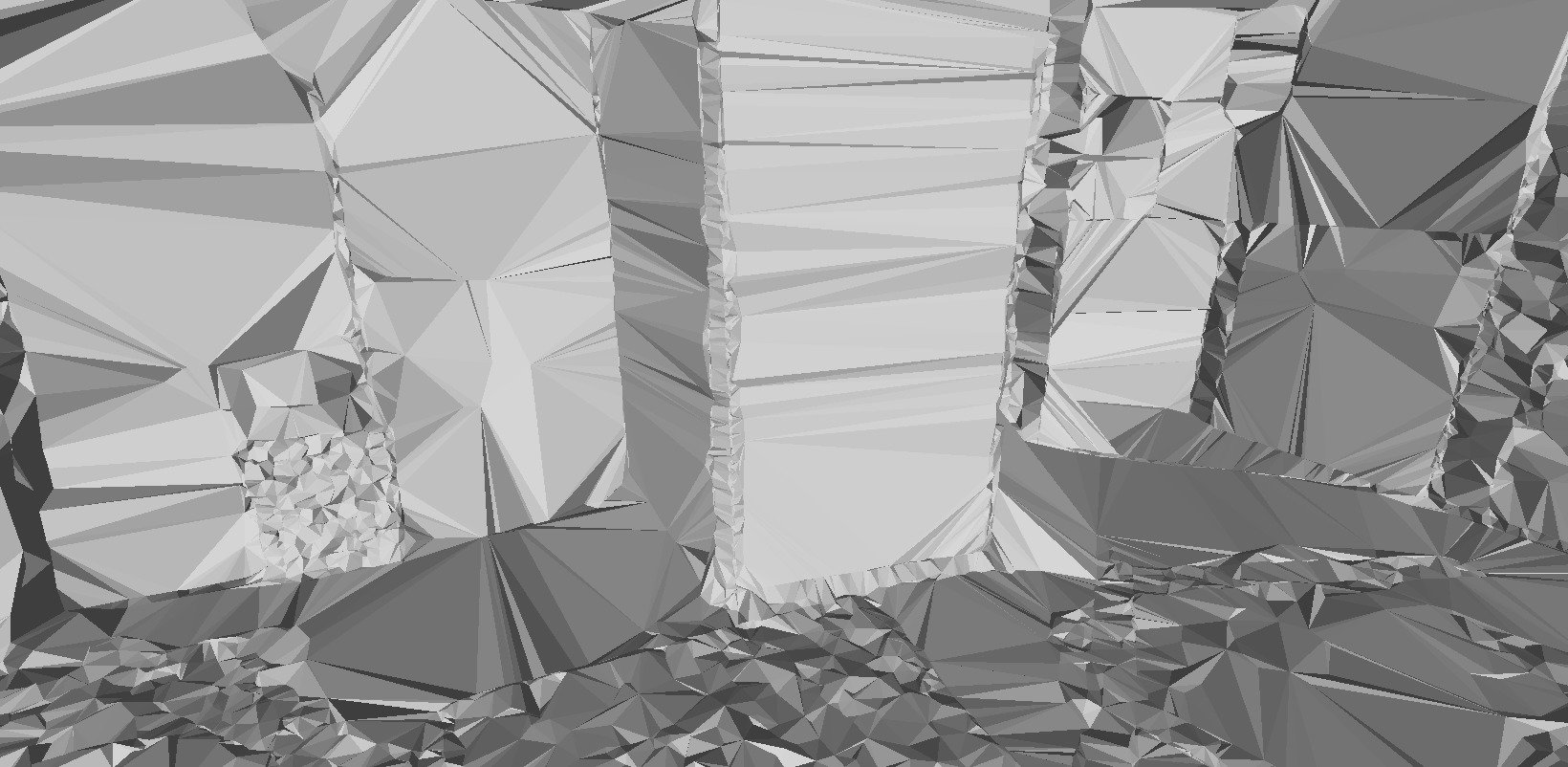}
		~
		\includegraphics[width=0.45\textwidth]{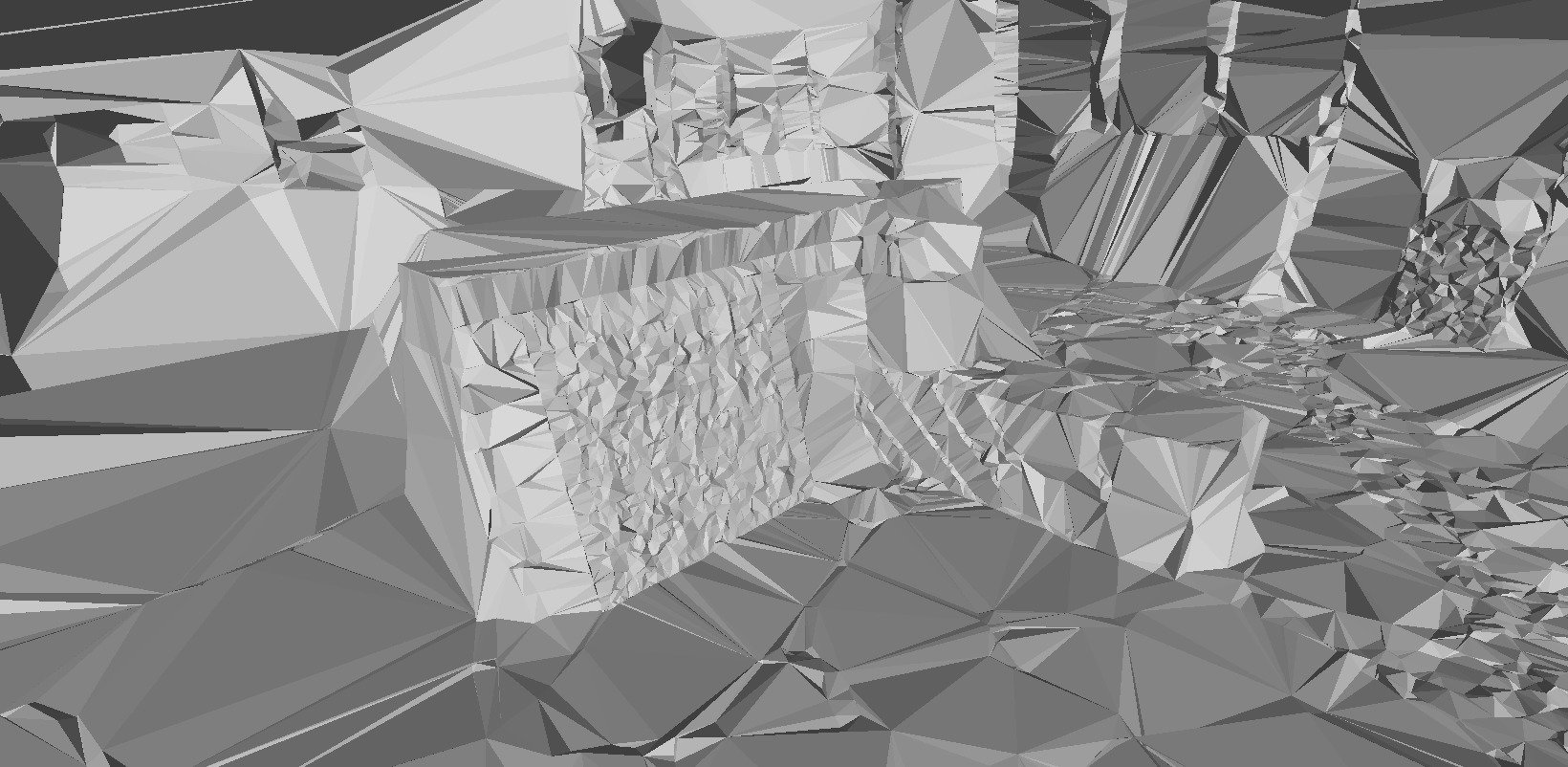}
		\end{minipage}
        }
		\caption{Reconstructed surface of sequence EuRoC MAV Vicon Room 101 in different views.}
		\label{fig:surface}
	\end{center}
\end{figure}

\subsection{Surface Reconstruction}

The resulting line segments of our method can be used to improve the quality of surface reconstruction.  
We integrate our method to the incremental space carving surface reconstruction algorithm presented in \cite{lovi2011incremental}. 
The algorithm incrementally reconstructs the surface by marking discretized 3D space as free or occupied using the visibility information of interest points.
We compare the reconstructed surface using the line segment endpoints from our proposed method versus using the map points of ORB-SLAM.
The results running on EuRoC Vicon Room 101 sequence are shown in Figure \ref{fig:surface}.
Thanks to the structural information and fewer outliers provided by our method, major structures in the room are much more obvious.

\section{Conclusion}

In this paper, we present an incremental 3D line segment based method that uses underlying structural information to simplify the semi-dense point cloud output by keyframe-based SLAM system. 
The main contribution lies in the novel edge aided 3D line segment extraction algorithm which solely relies on the image and the semi-dense depth map of individual keyframes. 
We show that the result of our method is accurate and can be used in incremental surface reconstruction to improve the quality of 3D surfaces.







%



\bibliographystyle{IEEEtran}
\bibliography{egbib}

\end{document}